%% file: main.tex
\documentclass[10pt,twocolumn,letterpaper]{article}

\usepackage[pagenumbers]{wacv} 

\usepackage{multirow} 

\definecolor{wacvblue}{rgb}{0.21,0.49,0.74}
\usepackage[pagebackref,breaklinks,colorlinks,allcolors=wacvblue]{hyperref}

\def\wacvPaperID{2023} 
\def\confName{WACV}
\def\confYear{2027}

\title{Heeding the Inner Voice: Aligning ControlNet Training via Intermediate Features Feedback}

\author{Nina Konovalova$^*$\\
MBZUAI, FusionBrain Lab\\
\and
Ibragim Idrisov$^*$\\
Lomonosov Moscow State University\\
\and
Kamil Garifullin\\
HSE University, FusionBrain Lab\\
\and
Timofei Sobornov\\
HSE University\\
\and
Aibek Alanov\\
HSE University, FusionBrain Lab\\
}

\begin{document}
\maketitle
{\let\thefootnote\relax\footnotetext{$^*$Equal contribution.}}
\input{sec/0_abstract}    
\input{sec/1_intro}

\input{sec/related}
\input{sec/method}
\input{sec/experiments}

\input{sec/ablations}
\input{sec/conclusion}\par
{
    \small
    \bibliographystyle{ieeenat_fullname}
    \bibliography{main}
}
\clearpage
\appendix
\input{sec/appendix}

\end{document}

%% file: sec/0_abstract.tex
\begin{abstract}
Despite significant progress in text-to-image diffusion models, achieving precise spatial control over generated outputs remains challenging. One of the popular approaches for this task is ControlNet, which introduces an auxiliary conditioning module into the architecture. To improve alignment of the generation image and control, ControlNet++ proposes a cycle consistency loss to refine correspondence between controls and outputs, but restricts its application to the final denoising steps, while the main structure is introduced at an early stage of generation. To address this issue, we suggest \textbf{InnerControl} -- a training strategy that enforces spatial consistency across a broad range of diffusion steps, including early high-noise ones. Specifically, we train lightweight control prediction probes — small convolutional networks -- to reconstruct input control signals (e.g., edges, depth) from intermediate diffusion features across denoising steps. We show empirically that such models can extract signals even from very noisy latents and utilize these models to generate pseudo ground truth controls during training. Suggested approach enables alignment loss that minimizes the difference between predicted and target condition over most of the diffusion trajectory. Our experiments demonstrate that our method improves control alignment, with perceptual fidelity preserved or improved in most settings and small FID trade-offs in others. Our approach can be applied to both UNet- and Transformer-based diffusion architectures and be integrated with established training techniques (e.g., ControlNet++) to achieve high performance across different condition methods such as edge, segmentation and depth conditions. The project page is available \href{https://controlgenai.github.io/InnerControl/}{here}.
\end{abstract}

%% file: sec/1_intro.tex
\section{Introduction}
\label{sec:intro}

Recent advances in diffusion models~\cite{ho2020denoising,song2020denoising,sohl2015deep, dhariwal2021diffusion} have significantly improved the quality and diversity of text-to-image (T2I) generation, enabling models to produce images that closely match input textual prompts~\cite{nichol2021glide, ramesh2021zero, rombach2022high, saharia2022photorealistic}. Despite this progress, achieving \textbf{precise spatial control} over generated images remains a key challenge~\cite{hu2023cocktail, ye2023ip, huang2023composer}. 

To address this issue, methods such as ControlNet~\cite{zhang2023adding} and T2I-Adapter~\cite{mou2024t2i} introduce conditional mechanisms to guide the generation process using control signals (e.g., edge maps, depth maps, segmentation masks). Follow-up work has aimed to improve ControlNet performance through architectural enhancements~\cite{zavadski2024controlnet}, unified conditioning ~\cite{qin2023unicontrol, zhao2023uni}, and efficient adaptation to new conditions~\cite{xu2024ctrlora}.  However, these approaches often suffer from \textbf{inconsistencies between the input control signals and the final generated output. }

Recent methods aim to improve control fidelity by introducing additional supervision during training. For example, ControlNet++~\cite{li2024controlnet++} reduces the discrepancy between the input control signal and the generated image by applying reward losses that penalize inconsistencies between the generated output and the extracted control signals (e.g., edges or depth). In contrast, CTRL-U~\cite{zhang2024ctrl} proposes an alternative approach based on uncertainty-aware reward modeling, which aims to mitigate the negative impact of inaccurate or noisy feedback from reward models. While both methods demonstrate improved control alignment, they primarily operate at \textbf{late denoising steps}, despite evidence that spatial structure emerges early in the diffusion process~\cite{chen2306beyond, baranchuk2021label}.

However, extending the reward losses to earlier steps leads to a significant decrease in the generated image quality, producing visible artifacts on the generated images. These poor results are probably caused by inefficient signal extraction in the early sampling steps, producing inaccurate signals for loss calculations. This analysis highlights a critical limitation of the suggested approaches, making them applicable only to the late generation steps.

To address temporal misalignment in prior methods, we introduce \textbf{InnerControl} -- a novel training strategy that enforces consistency between input controls (e.g., edges, depth, segmentation) and signals extracted from \textbf{intermediate diffusion features} over a broad, task-dependent range of denoising steps (including early high-noise ones).
Our approach is motivated by recent findings that demonstrate the utility of diffusion features for vision tasks such as depth estimation, semantic segmentation, and classification~\cite{baranchuk2021label, hedlin2023unsupervised, namekata2024emerdiff}. Building on this, we propose to use lightweight networks to extract control signals directly from diffusion model features.

Drawing inspiration from Readout Guidance~\cite{luo2024readout} and DiTF~\cite{gan2025unleashing}, we employ timestep-conditioned estimation models to introduce an additional alignment penalty during ControlNet training. We keep the formulation unchanged across backbone families. Our findings indicate that InnerControl enhances the previous reward-based training approach, improving control alignment; perceptual quality is preserved or improved in most settings, with small FID changes in others (Tables~\ref{tab:unified_control}--\ref{tab:modern_depth}).

Our core contributions are:

\begin{itemize}
    \item \textbf{Early-stage control alignment:} we propose a novel training objective that enforces consistency between the input control signal (e.g., edge and depth maps, etc.)  and signals extracted from intermediate diffusion features across broad task-specific timestep ranges, including early stages where structural content begins to emerge.
    \item \textbf{Improved controllability} : unlike reward-based losses, which degrade image quality when applied to early denoising steps, our alignment loss consistently improves controllability across the evaluated tasks and backbones. Its effect on perceptual quality depends on the setting: FID is preserved or improved on several pipelines, while small degradations appear in others (e.g., LineArt FID; SDXL/FLUX under reward training).
    \item \textbf{Model and pipeline generality:} we validate proposed alignment objective across UNet-based (SD1.5, SDXL) and Transformer-based models (SD3.5, FLUX), and within standard ControlNet-style training pipelines. Results indicate consistent improvements in controllability without modifications to the core loss design.
\end{itemize}

%% file: sec/related.tex
\section{Related Work}
\subsection{Controllable Text-to-Image Diffusion Models}
Diffusion models \cite{ho2020denoising,song2020denoising,sohl2015deep, dhariwal2021diffusion} have achieved remarkable success in generating high-quality, diverse images conditioned on text prompts \cite{nichol2021glide, ramesh2021zero, rombach2022high, saharia2022photorealistic, balaji2022ediff, ramesh2022hierarchical}.
However, traditional approaches rely solely on textual guidance, limiting precise spatial control over generated outputs. To address this, several methods introduce spatial control signals without retraining the entire diffusion pipeline. ControlNet~\cite{zhang2023adding} augments pretrained diffusion models with a duplicate encoder and zero-convolution layers, enabling stable training and alignment with diverse spatial conditions (e.g., edges, depth, segmentation masks). Similarly, T2I-Adapter~\cite{mou2024t2i} employs lightweight adapter modules to bridge internal text-to-image representations with external control inputs. Subsequent works refine these designs for greater efficiency~\cite{zavadski2024controlnet,cao2025relactrl}, extend them into unified frameworks supporting multiple control types~\cite{zhao2023uni,qin2023unicontrol}, or improve adaptability to novel control signals~\cite{xu2024ctrlora} and advanced backbones~\cite{lin2024ctrl,ran2024x}. Despite these advances, ensuring consistent alignment between generated outputs and conditioning signals remains challenging. 

To address this issue, ControlNet++~\cite{li2024controlnet++} introduces an additional reward loss for ControlNet that enhances controllable generation by explicitly optimizing pixel-level cycle consistency between generated images and conditional inputs. Ctrl-U~\cite{zhang2024ctrl}, on the other hand, introduces uncertainty-aware reward modeling to regularize reward fine-tuning through consistency construction. Specifically, rewards with lower uncertainty are assigned higher loss weights, while those with higher uncertainty receive reduced weights to accommodate greater variability. However, both approaches primarily target late-stage alignment due to the suggested one-step prediction strategy for signal estimation, thereby neglecting the earlier phases of generation. Meanwhile, prior studies indicate that the main structure emerges during the early stages of generation~\cite{chen2306beyond}. This makes consistency over a wide portion of the generation trajectory essential for preserving fidelity to input conditions. Our approach addresses this gap by enforcing alignment over task-specific ranges that include early high-noise steps. Unlike Readout Guidance, which uses readout heads for inference-time guidance, we use them only as a training-time loss. Applying them at inference did not improve controllability and increased sampling time (Supp.~\ref{discussion}).

\subsection{Diffusion Model Representation}
Pretrained text-to-image diffusion models have proven highly effective at extracting semantically rich representations from their internal features, supporting a wide range of discriminative tasks such as segmentation, semantic correspondence, classification, detection, and depth estimation~\cite{fundel2025distillation,hedlin2023unsupervised,clark2023text,yang2023diffusion,xiang2023denoising}. Several studies have analyzed the quality of UNet features across different denoising steps for downstream vision tasks~\cite{baranchuk2021label,chen2306beyond}.

Recent efforts extend this line of work by aggregating features across layers and denoising steps to improve performance, with applications in segmentation~\cite{tang2022daam,namekata2024emerdiff,stracke2024cleandift}, semantic correspondence~\cite{tang2023emergent,hedlin2023unsupervised,stracke2024cleandift}, classification~\cite{li2023your,stracke2024cleandift}, detection~\cite{chen2023diffusiondet,stracke2024cleandift}, and depth estimation~\cite{stracke2024cleandift}. 

However, most existing approaches rely on aggregated features across denoising steps, which may limit their ability to capture task-specific information at each individual stage. To address this limitation, Luo et al.~\cite{luo2024readout} enhance the Diffusion Hyperfeatures framework~\cite{luo2023diffusion} by introducing additional timesteps conditioning. Similarly, for Transformer-based diffusion, feature extraction approach based on post-normalization was suggested in work DiTF~\cite{gan2025unleashing}.


\begin{figure*}[!ht]
\centering
\includegraphics[width=0.9\linewidth]{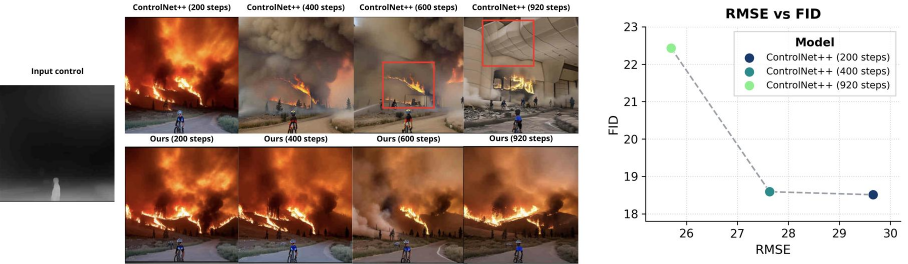} 
\caption{
Visualizing the trade-off between control consistency (RMSE) and image fidelity (FID) when extending reward losses to early denoising stages. \textit{Left:} 
Generated samples illustrate visual artifacts when applying only reward loss (\textit{top}) compared to reward plus alignment loss (\textit{bottom}) during early diffusion steps. \textit{Right:} Quantitative analysis shows the trade-off between control precision (RMSE ↓) and image fidelity (FID ↓), highlighting their inverse relationship.}

\label{fig:rewarding_ablation}
\vspace{-10pt}
\end{figure*}

%% file: sec/method.tex
\section{Preliminaries}
In this section, we introduce the background of diffusion models and spatially controllable generation, followed by an analysis of cycle consistency losses suggested in ControlNet++~\cite{li2024controlnet++}. 

\subsection{Controllable generation}
Diffusion models~\cite{ho2020denoising,song2020denoising} are a class of generative models that synthesize data by iteratively denoising random noise through a learned reverse process.
 
The forward process defines a sequence of noise-adding steps that transform the data into isotropic Gaussian noise over $T$ steps.
\begin{equation}
    q(x_t | x_{t-1}) = \mathcal{N}\left(x_t; \sqrt{\alpha_t} x_{t-1}, (1 - \alpha_t)\mathbf{I}\right)
\end{equation}
where $ \alpha_t \in (0, 1) $ is a fixed variance schedule. The reverse process  learns to invert this process using a neural network that iteratively denoises samples:
\begin{equation}
    p_\theta(x_{t-1} | x_t) = \mathcal{N}\left(x_{t-1}; \mu_\theta(x_t, t), \sigma_t^2\right),
\end{equation}
where $ \mu_\theta(\cdot) $ is a learnable function that approximates the mean of the true posterior.

The standard training objective aims to minimize the noise prediction error:
\begin{equation}
    \mathcal{L}_{\text{diff}} = \mathbb{E}_{x_0, \epsilon, t} \left[ \left\| \epsilon - \epsilon_\theta(x_t, t) \right\|^2 \right]
\end{equation}

where $x_t$ is a noisy sample from timestep $t$, $\epsilon \sim \mathcal{N}(0,I)$ and $ \epsilon_\theta(x_t, t)$ -- learned approximation using another parametrization.

In cases where additional control is required, such as text prompt conditioning $c_{txt}$ and spatial control $c_{spat}$ (e.g., depth maps or edges), the objective can be expressed as:
\begin{equation}
    \mathcal{L}_{\text{diff}} = \mathbb{E}_{x_0, \epsilon, t, c_{txt},c_{spat}} \left[ \left\| \epsilon - \epsilon_\theta(x_t, t, c_{spatial}, c_{txt}) \right\|^2 \right]
    \label{eq:diffusion_loss}
\end{equation}

\subsection{ControlNet++}
ControlNet~\cite{zhang2023adding} is one of the leading methods that utilizes a pretrained text-to-image diffusion model for controllable generation with additional spatial control. While ControlNet is trained using standard diffusion loss defined in Eq.~\ref{eq:diffusion_loss}, it suffers from inconsistencies between the final predictions and the input controls. To mitigate this issue, ControlNet++ introduces a cycle consistency loss that leverages a discriminative reward model.

Specifically, the method minimizes the discrepancy between the input control $c_{spatial}$ and the corresponding condition $\hat{c}_{spatial}$ extracted from the generated image by the reward model $\mathbb{D}$, where $\hat{c}_{spatial} = \mathbb{D}(x_0)$ and $x_0$ denotes the generated image. Since diffusion models sample timesteps $t \in [999, 0]$ to simulate the denoising process, computing rewards across the full trajectory would require prohibitive gradient accumulation. To address this, ControlNet++ approximates $x_0$ from a noisy sample $x_t$ using a single-step generation process:
\begin{equation}
\begin{split}
    x_0 &\approx x'_0 = \mathbb{G}(c_{\text{spat}}, c_{\text{txt}}, x_t, t) = \\
        &= \frac{x_t - \sqrt{1 - \alpha_t} \, \epsilon_\theta (x'_t, c_{\text{spatial}}, c_{\text{txt}}, t)}{\sqrt{\alpha_t}}
\end{split}
\end{equation}

where $\epsilon_{\theta}(\cdot)$ denotes the network's noise prediction and $G(\cdot)$ represents the single-step denoising operation.  
The resulting approximation $x'_0$ can then be used for reward fine-tuning:
\begin{equation}
\mathcal{L}_{\text{reward}} = \mathcal{L}\left(c_{spatial}, \mathbb{D}\left[\mathbb{G}(c_{spatial}, c_{txt}, x_t, t)\right]\right)
\label{eq:reward}
\end{equation}

Due to single-step sampling, the authors suggest applying their rewarding loss only on the last $200$ steps ($t \in [0, 200]$) of diffusion trajectory sampling.

\section{Method}

In this section, we discuss the main limitations of ControlNet++ and introduce our proposed training approach -- \textbf{InnerControl} -- designed to address these issues.

\subsection{Motivation}
As noted earlier, ControlNet++~\cite{li2024controlnet++} applies the reward loss only to the final 200 denoising steps ($t \in [0, 200]$) due to its reliance on a single-step prediction strategy. To better understand this limitation, we analyze the trade-off between control consistency and image fidelity when extending the reward loss to earlier denoising stages. Specifically, we train a ControlNet model conditioned on depth maps, applying $\mathcal{L}_{\text{reward}}$ loss over different ranges of denoising steps. Alignment is measured using RMSE, while perceptual quality is evaluated with FID.

Our experiments show that extending the reward loss to earlier steps improves control alignment, as reflected in lower RMSE (Fig.~\ref{fig:rewarding_ablation}). While RMSE improves, the quality of generated images is significantly decreased, leading to an increase in FID (Fig.~\ref{fig:rewarding_ablation}, \textit{Right}). The perceptual metrics degradation can also be observed in the images, where artifacts appear as the reward loss is extended to earlier steps. 

As shown in Fig.~\ref{fig:rewarding_ablation}, applying the reward loss up to $400$ denoising steps produces good-quality images, but extending it to $600$ steps introduces small artifacts, and using it across nearly all steps (e.g., $920$) results in severe distortions such as unexpected lines and irregular edges.

\begin{figure}[!htb]
    \centering
    \includegraphics[width=0.95\columnwidth]{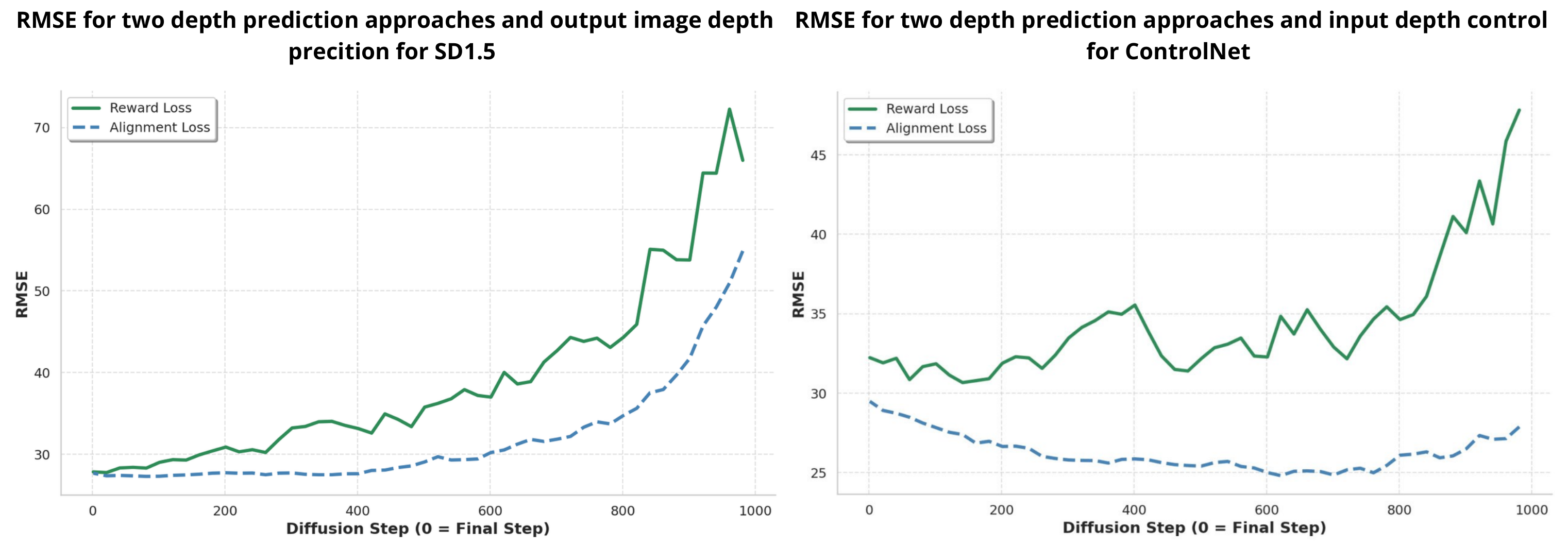}
    \caption{\textit{Left:} RMSE between final-image depth and DPT prediction for single-step predicted images (green) and intermediate-feature depth estimates (blue) in SD1.5. \textit{Right:} RMSE between control depth and DPT prediction for single-step predicted images (green) and intermediate-feature depth estimates (blue) in ControlNet.}     
    \label{fig:losses}
\vspace{-8pt}
\end{figure}
\raggedbottom

This degradation arises from the poor quality of single-step predictions at highly noisy timesteps. In early denoising stages, the single-step image predictions are extremely blurry and fall outside the domain expected by pretrained depth estimators. Fig.~\ref{fig:blurry} illustrates this effect: from around 400 steps onward, single-step predictions become increasingly blurry, leading to unreliable DPT depth estimates. This is further supported in Fig.~\ref{fig:losses}, where RMSE increases substantially at early steps (green line) for both SD1.5 and ControlNet generations. Consequently, the extracted depth maps are inaccurate and misaligned with ground-truth depth map of an image, propagating errors during training.

\subsection{Alignment on early steps}


\begin{figure}[!htb]
\centering
    \centering
    \includegraphics[width=0.95\columnwidth]{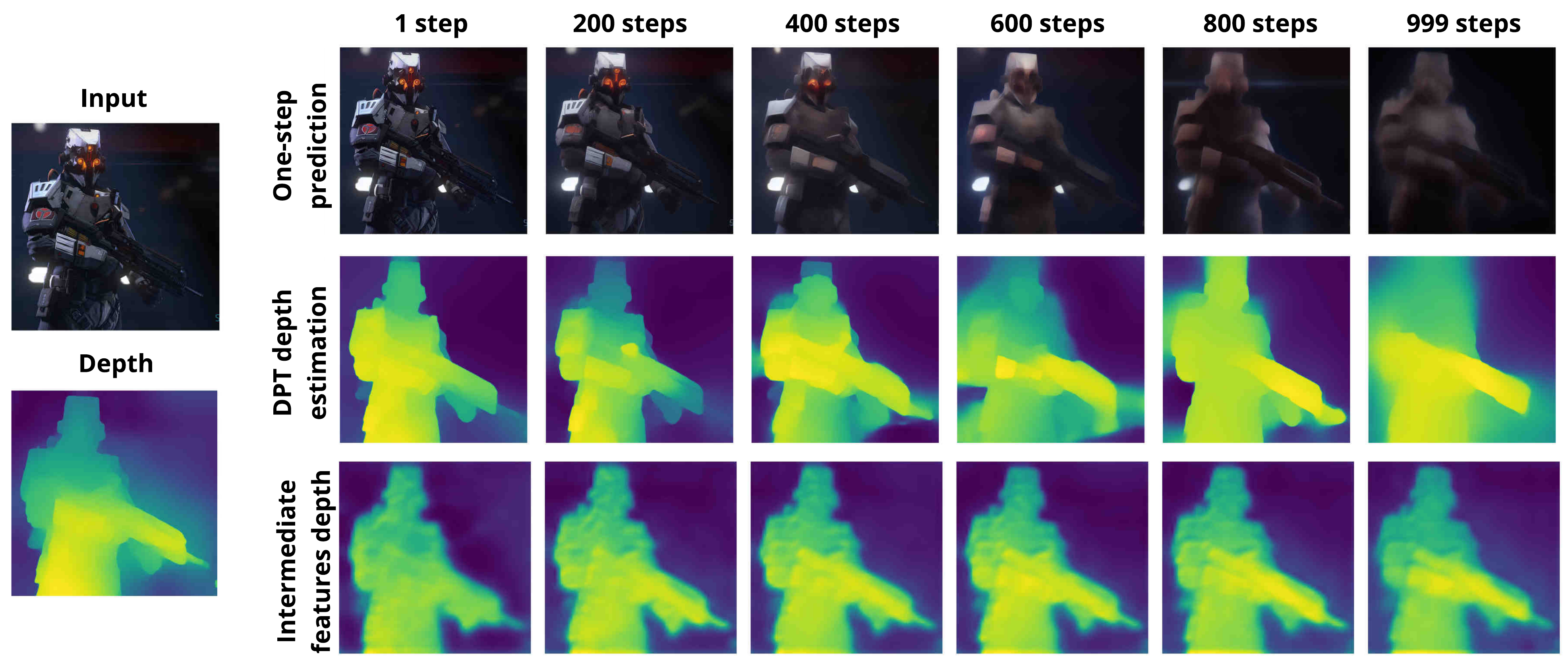}
     \caption{One-step predictions (\textit{top}) at increasing noise levels, with corresponding depth maps from DPT~\cite{ranftl2021vision} (\textit{middle}) and intermediate UNet features (\textit{bottom}).}
    \label{fig:blurry}

\vspace{-10pt}
\end{figure}

While standard discriminative models such as DPT~\cite{ranftl2021vision} struggle to extract reliable control signals from blurry images, prior work shows that intermediate diffusion features already encode spatial structure at early denoising stages~\cite{chen2306beyond}. Motivated by this observation, we train a lightweight convolutional probe $\mathbb{H}(\cdot,t)$ to predict control signals directly from intermediate features instead of image-space one-step reconstructions.

Following Readout Guidance~\cite{luo2024readout} (and DiTF~\cite{gan2025unleashing} for Transformer-based model evaluation), we keep the probe architecture timestep-conditioned. We validate feature-space estimation by comparing probe predictions with DPT depth estimates~\cite{ranftl2021vision} (Fig.~\ref{fig:blurry}).

Our results demonstrate that $\mathbb{H}(\cdot,t)$ predicts results more aligned with the final depth prediction for SD1.5, especially at the early stage of generation Fig.~\ref{fig:losses}. Additionally, $\mathbb{H}(\cdot,t)$ proves to be more stable for ControlNet generation, maintaining consistent signal estimation across a wide range of denoising steps (Fig.~\ref{fig:losses}, \textit{Right}). These findings indicate that intermediate features provide more robust signal predictions, enabling accurate control signal estimation even in high-noise regimes. By leveraging $\mathbb{H}(\cdot,t)$ to enforce alignment at early stages of generation,  we introduce \textbf{InnerControl} -- the method that suggests the new training objective that addresses the misalignment that appeared in the previous approaches:  
\begin{equation}
\mathcal{L}_{\text{alignment}} = \mathcal{L}\left(c_{\text{spatial}}, \, H_t\right)
    \label{eq:alignment}
\end{equation}
where $H_t = \mathbb{H}\left( \text{ControlNet}(c_{\text{spatial}}, c_{\text{txt}}, x_t, t),\, t \right)$ is the lightweight convolutional estimator applied to ControlNet features at timestep $t$. The loss is MSE for depth and edges and cross-entropy for segmentation. The details about $\mathbb{H}(\cdot,t)$ training can be found in Appendix.

This loss is applied during training to improve control alignment. The alignment block is illustrated in Fig.~\ref{fig:pipeline}.

For the final training objective, we use a weighted combination of the standard diffusion loss~\ref{eq:diffusion_loss}, reward loss at the late denoising stage~\ref{eq:reward}, and the additional alignment loss~\ref{eq:alignment}:
\begin{equation}
\mathcal{L}_{\text{training}} = \mathcal{L}_{\text{diffusion}} + \alpha \cdot \mathcal{L}_{\text{reward}} + \beta \cdot \mathcal{L}_{\text{alignment}} 
\label{eq:full_loss}
\end{equation}

Additional loss penalizes discrepancies between $c_{spatial}$ and the probe prediction ($H_t$) at each timestep $t$, enforcing spatial alignment over the selected timestep ranges and resulting in improved control correspondence and image fidelity, without visible artifacts compared to reward loss at early steps (Fig.~\ref{fig:rewarding_ablation}).


\textbf{Effect of Probe Alignment on Generated Images.} {The probe $\mathbb{H}(\cdot,t)$ is used only during training and discarded at inference, while all reported controllability metrics are computed on final decoded images using external task evaluators. Moreover, the probe is trained on features of the base diffusion model, independently of ControlNet fine-tuning, and remains frozen during alignment training (Appendix). With the reward loss disabled, alignment alone improves final-image depth RMSE on SD1.5, SDXL, and FLUX (Table~\ref{tab:combined_ablation_narrow}), showing that improved probe-space alignment transfers to the structure of generated images rather than merely improving an internal representation. Qualitatively, the intermediate readouts of InnerControl remain aligned with the input control across denoising timesteps, whereas those of a standard ControlNet diverge (Appendix Fig.~\ref{fig:diff}).}

%% file: sec/experiments.tex
\begin{figure}[!htb]
\centering
    \includegraphics[width=0.95\columnwidth]{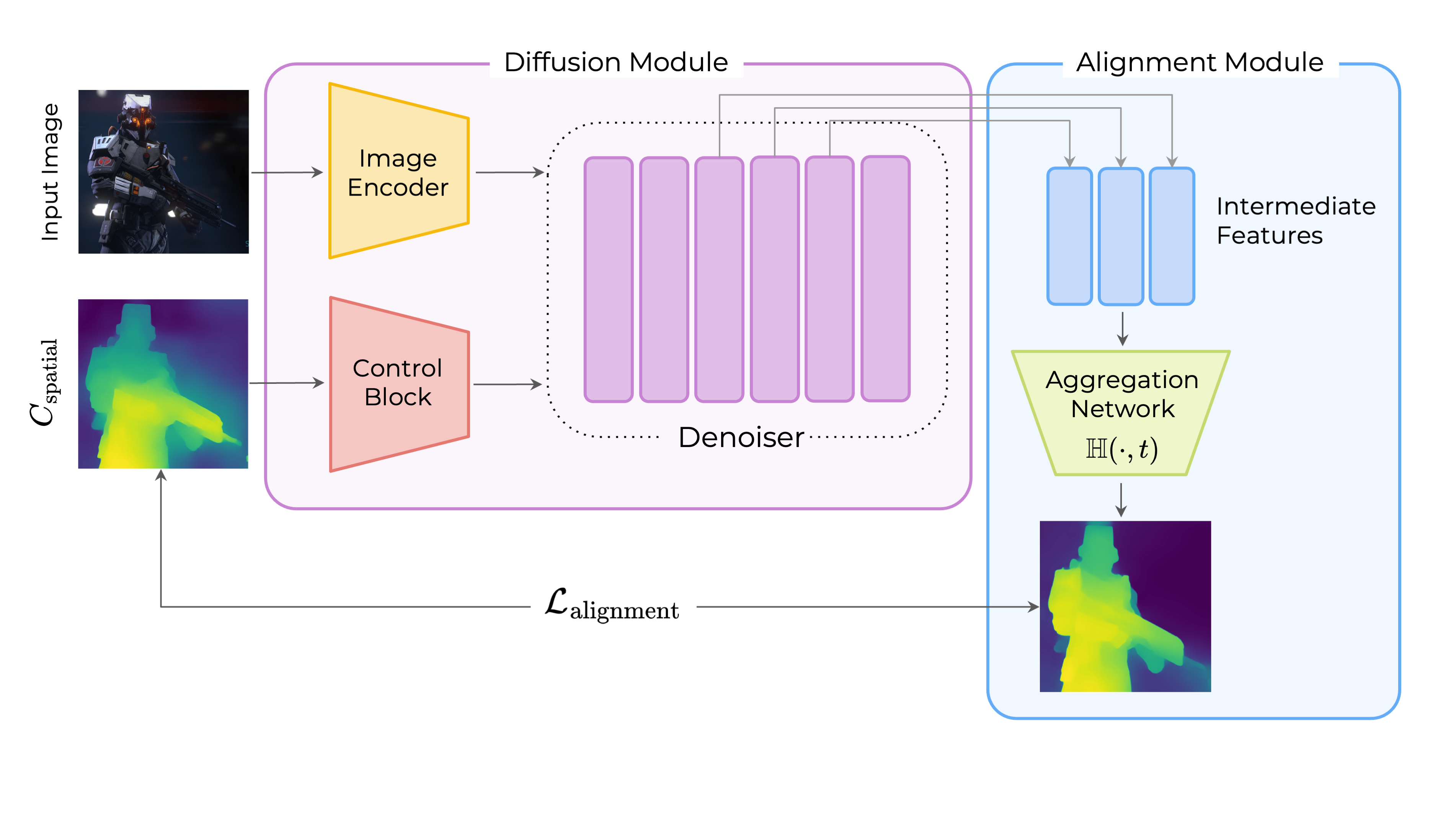}
    \vspace{-10pt}
    \caption{\textbf{Pipeline overview.} Schematic illustration of our \textbf{InnerControl} framework, highlighting the integration of the alignment loss. The key difference from ControlNet++ is the \textbf{Alignment module}, which processes intermediate features extracted from the diffusion denoiser (UNet decoder features and Transformer block features). These features are passed through a lightweight aggregation network to predict spatial control signals (e.g., depth or edge maps), which are then compared with the input control $c_{spat}$ to enforce consistency at each denoising step.}
    \label{fig:pipeline}
\vspace{-10pt}
\end{figure}
\raggedbottom

\section{Experiments}

\subsection{Experiment setup}\label{sec:exp_details}

\begin{table*}[h!]
    \centering
    \caption{Unified comparison on the MultiGen‑20M benchmark (segmentation on ADE20K). Controllability is evaluated using SSIM ($\uparrow$) for HED/LineArt, RMSE ($\downarrow$) for depth, and mIoU ($\uparrow$) for segmentation; fidelity is measured by FID ($\downarrow$). Numbers for earlier baselines are taken from prior publications; ControlNet++, Ctrl‑U, and InnerControl are re-trained by us under identical settings (Sec.~\ref{sec:exp_details}).}
    
    \label{tab:unified_control}
    \resizebox{\linewidth}{!}{
    \begin{tabular}{@{}l c c c c c c c c c@{}}
        \toprule
        \multirow{2}{*}{\textbf{Method}} &
        \multirow{2}{*}{\textbf{T2I Model}} &
        \multicolumn{2}{c}{\textbf{Hed Edge}} &
        \multicolumn{2}{c}{\textbf{LineArt Edge}} &
        \multicolumn{2}{c}{\textbf{Depth Map}} &
        \multicolumn{2}{c}{\textbf{Segmentation}} \\
        \cmidrule(lr){3-4}\cmidrule(lr){5-6}\cmidrule(lr){7-8}\cmidrule(lr){9-10}
        & & SSIM $\uparrow$ & FID $\downarrow$
          & SSIM $\uparrow$ & FID $\downarrow$
          & RMSE $\downarrow$ & FID $\downarrow$
          & mIoU $\uparrow$ & FID $\downarrow$ \\
        \midrule
        \multicolumn{10}{l}{\textit{Guidance scale = 7.5}}\\
        \midrule
        ControlNet              & SDXL  & —           & —            & —           & —            & 40.00 & —            & —           & — \\
        T2I‑Adapter             & SDXL  & —           & —            & 0.639      & —            & 39.75 & —            & —           & — \\
        T2I‑Adapter             & SD1.5 & —           & —            & —           & —            & 48.40 & 22.52        & —           & — \\
        Gligen                  & SD1.4 & 0.563      & —            & —           & —            & 38.83 & 18.36        & —           & — \\
        Uni‑ControlNet          & SD1.5 & 0.691      & 17.1        & —           & —            & 40.65 & 20.27        & —           & — \\
        UniControl              & SD1.5 & 0.797      & 16.0        & —           & —            & 39.18 & 18.66        & —           & — \\
        ControlNet              & SD1.5 & 0.762      & 15.4        & 0.705     & 17.4        & 35.90 & \textbf{17.76}        & 32.60           & 41.1 \\
        ControlNet++   & SD1.5 & \underline{0.822}           & \textbf{13.0}            & \underline{0.840}      & \underline{13.2}            & 27.63 & 18.59 & \underline{38.08}          & 39.04 \\
        Ctrl‑U         & SD1.5 & 0.820 & \underline{13.2} & 0.810 & \textbf{12.5} & \underline{26.50} & 18.67        & 36.95           & \textbf{35.00}\\
        \textbf{InnerControl} (Ours) & SD1.5 & \textbf{0.826} & \textbf{13.0} & \textbf{0.850} & 13.5 & \textbf{26.09} & \underline{18.29}        & \textbf{40.22}           & \underline{37.65} \\
        \bottomrule
    \end{tabular}
    }
\vspace{-6pt}
\end{table*}

\textbf{Datasets.}
Our method was evaluated on multiple datasets, corresponding to a specific task. We evaluated LineArt and HED conditioning on the MultiGen-20M dataset~\cite{qin2023unicontrol}, a large-scale synthetic dataset containing paired images and control signals. For depth estimation, we utilized the corresponding MultiGen-20M depth dataset, which provides precomputed depth maps generated using standard monocular depth estimation techniques. Segmentation experiments were conducted on ADE20K~\cite{zhou2019semantic,zhou2017scene}.\\
\textbf{Implementation details.} 
To ensure a fair comparison, we trained our model, ControlNet++~\cite{li2024controlnet++} and Ctrl-U~\cite{zhang2024ctrl} under identical experimental settings. For each type of control, we first finetuned the pretrained ControlNet model using the AdamW optimizer with a learning rate of $10^{-5}$.

For UNet-based models we run this stage for $5$k iterations for segmentation and $10$k iterations for all other tasks. After finetuning, we continue training for another $5$k iterations for segmentation and $10$k iterations for the remaining tasks, using the same optimization settings and the proposed loss~\ref{eq:full_loss}.
Specifically, the proposed $\mathcal{L}_{\text{alignment}}$ loss was applied over different ranges of diffusion steps: $[0, 920]$ for depth, $[0, 800]$ for HED, $[0, 700]$ for LineArt, and $[450, 980]$ for segmentation. The reward loss was applied over 200 steps for the LineArt and segmentation tasks, and over 400 steps for the depth and HED tasks.

For Transformer-based models, we follow the same training objective and focus only on the depth-control task due to limited resources. To optimize computational efficiency, we reduce the number of iterations for both ControlNet tuning and the subsequent alignment phase. Specifically, we cap these stages at 5,000 iterations for SD3.5 and 2,000 iterations for FLUX. Furthermore, the alignment loss is applied selectively within the $[400, 920]$ diffusion timestep only.
Following DiTF~\cite{gan2025unleashing}, we use a compact subset of post-normalization features for the alignment probe to improve representation quality while keeping the rest of the training pipeline unchanged.

All experiments used $512 \times 512$ images with an effective batch size of $256$. See the supplementary material for detailed model settings.\\
\textbf{Alignment model.}
Following this approach, we develop an aggregation network that extracts features from the UNet decoder and employs bottleneck layers to standardize channel counts, which are then combined using a learned weighted sum. The model is conditioned on each diffusion step via pretrained timestep embeddings. To improve edge estimation, we introduce a 2D transposed convolution operator in the output head. For Transformer backbone we utilize the same probe architecture while extracting normalization modulated features according to DiTF~\cite{gan2025unleashing}. Detailed specifications of the alignment model are provided in the Appendix, where we also demonstrate that a unified estimation alignment model can be trained for subsequent use in ControlNet training.\\
\textbf{Baselines.}
We compare our method with T2I-Adapter~\cite{mou2024t2i}, ControlNet v1.1~\cite{zhang2023adding}, GLIGEN~\cite{li2023gligen}, Uni-ControlNet~\cite{zhao2023uni}, UniControl~\cite{qin2023unicontrol}, ControlNet++~\cite{li2024controlnet++}, and Ctrl-U~\cite{zhang2024ctrl}.

Most of these methods are based on SD1.5 for text-to-image generation, but we additionally include several models based on SDXL~\cite{podell2023sdxl}: ControlNet-SDXL and T2I-Adapter-SDXL, following metrics presented in Ctrl-U~\cite{zhang2024ctrl}. For a fair comparison, all models were evaluated under identical image conditions and  prompts, using guidance scale $7.5$ for UNet-based and $4.5$ for Transformer-based models. \\
\textbf{Metrics and evaluation.}
We evaluate alignment fidelity using task-specific metrics: Structural Similarity Index (SSIM) between generated edges and input control signals, Root Mean Squared Error (RMSE) between predicted and ground-truth depth maps and mIoU for the segmentation task. All metrics were computed on the $512 \times 512$ images to ensure consistency. To reduce stochastic variance, we generated $4$ independent sample batches with different random seeds and reported the mean metrics. More information about evaluation models can be found in Appendix.

\subsection{Experimental results}
\textbf{Comparison of Controllability.} We summarize the results on control alignment quality in Table~\ref{tab:unified_control}. Our method achieves notable improvements over the baselines for both depth and segmentation estimation: RMSE is reduced by $5.6\%$ and the mean Intersection over Union (mIoU) is improved by $5.6\%$ compared to ControlNet++. These results highlight stronger alignment with the control signals, particularly under high guidance intensity. Furthermore, for edge-based control tasks (LineArt, HED), our approach outperforms both ControlNet++ and Ctrl-U in terms of SSIM. \\
\textbf{Comparison of Image Quality.}
To evaluate the perceptual quality of generated images, we report the Fréchet Inception Distance (FID)~\cite{heusel2017gans} for all evaluated methods at a guidance scale of $7.5$ for UNet-based models over all different control tasks. As shown in Table~\ref{tab:unified_control}, our method achieves the best controllability metrics across all four tasks. Relative to ControlNet++, FID is unchanged on HED, improves on depth and segmentation, and is slightly worse on LineArt. Relative to Ctrl-U, our method provides stronger controllability while trading off FID on LineArt and segmentation. These results position InnerControl at a different point on the controllability–fidelity frontier rather than showing uniform improvements in both metrics. Because the alignment loss is independent of the reward formulation, it can also be added on top of Ctrl-U. We report this combination in Table~\ref{tab:ctrlu_comparison} and find it improves controllability there as well. 


\begin{table}[h!]
    \centering
    \caption{Depth-control results for modern diffusion backbones on MultiGen-20M.}
    \label{tab:modern_depth}
    \begin{tabular}{l c c c}
        \toprule
        \textbf{Method} & \textbf{Model} & \textbf{RMSE $\downarrow$} & \textbf{FID $\downarrow$} \\
        \midrule
        \multicolumn{4}{l}{\textit{Guidance scale = 7.5}}\\
        \midrule
        ControlNet++        & SDXL & \underline{30.24} & \textbf{17.00} \\
        InnerControl (Ours) & SDXL & \textbf{28.31} & \underline{17.10} \\
        \midrule
        \multicolumn{4}{l}{\textit{Guidance scale = 4.5}}\\
        \midrule
        ControlNet++        & SD 3.5 & \underline{31.07} & \textbf{14.24} \\
        InnerControl (Ours) & SD 3.5 & \textbf{30.31} & \underline{14.34} \\
        \midrule
        ControlNet++        & FLUX.1 & \underline{29.22} & \underline{23.61} \\
        InnerControl (Ours) & FLUX.1 & \textbf{27.91} & \textbf{22.54} \\
        \bottomrule
    \end{tabular}
\vspace{-6pt}
\end{table}

\begin{figure*}[h!]
\centering
    \centering
    \includegraphics[width=0.85\linewidth]{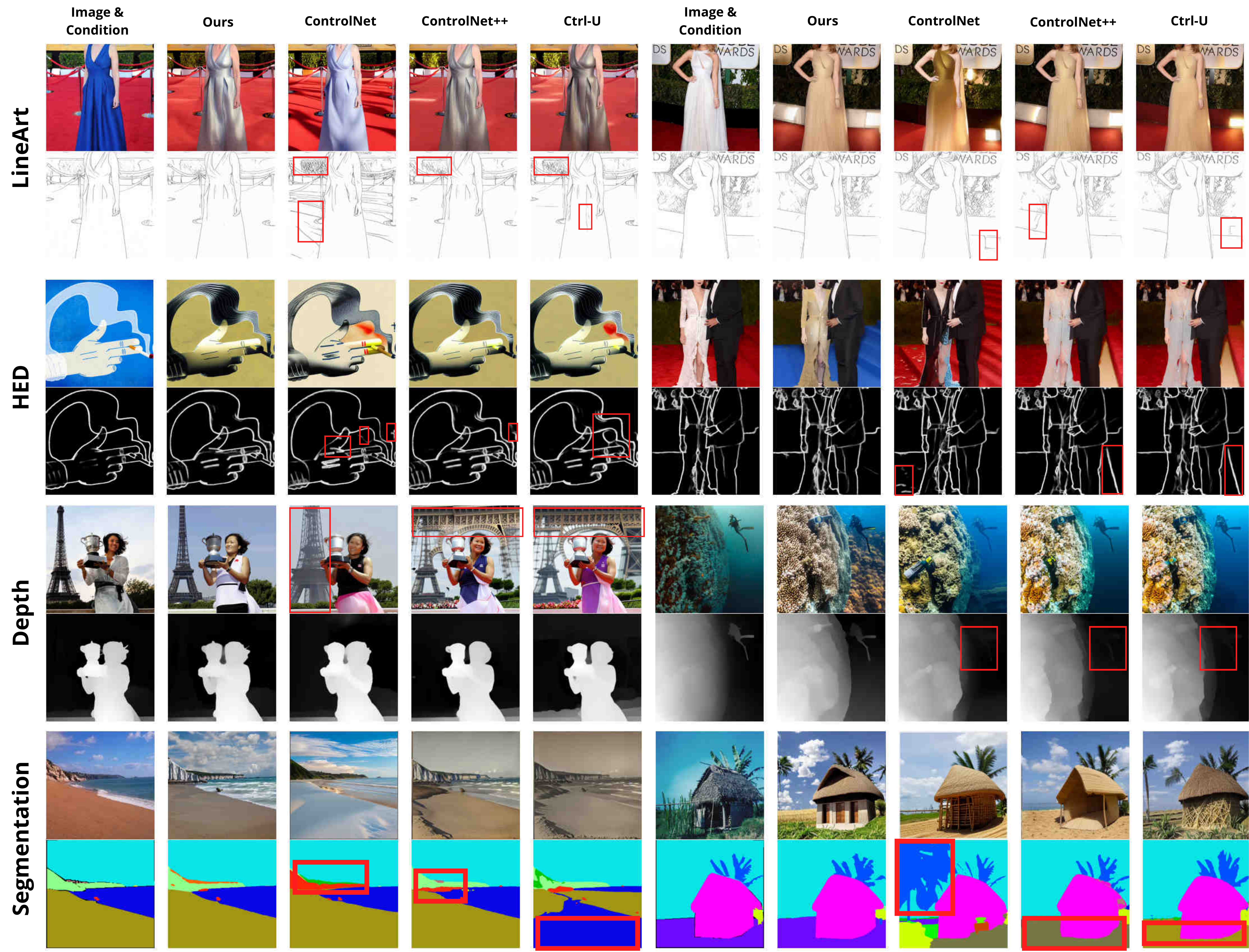}
    \caption{\textbf{Qualitative Comparison with Baselines:} Results for LineArt (top), HED and Depth (middle), and segmentation (bottom), using identical prompts and a guidance scale of $7.5$. Our method better aligns outputs with input controls than competing approaches.}
    \label{fig:results}  
\vspace{-10pt}
\end{figure*}
\textbf{Qualitative Analysis.} We present a qualitative comparison of image generations, showing side-by-side results from our method, ControlNet~\cite{zhang2023adding}, ControlNet++~\cite{li2024controlnet++}, and Ctrl-U~\cite{zhang2024ctrl} under identical prompts and control signals (i.e., depth maps, segmentation maps, HED edges, and LineArt edges) at a guidance scale of $7.5$. As highlighted in Fig.~\ref{fig:results}, we observed misalignments between the input conditions and the generated results of the competing models. Specifically, ControlNet produced noisy edges for both LineArt and HED control tasks, while ControlNet++ and Ctrl-U generated images with noticeable artifacts and inconsistent object distances when using depth control. Furthermore, all competitor models exhibited inaccurate object placement relative to the input segmentation mask during the segmentation task. A user study confirming these preferences is reported in the Appendix. \\
\textbf{Comparison on Modern Models.} We report additional results on the depth-control task for larger diffusion backbones (SDXL, SD3.5, and FLUX) in Table~\ref{tab:modern_depth}, showing overall better controllability; relative to ControlNet++, FID improves on FLUX and changes only marginally on SDXL/SD3.5. 
While we focused our evaluation on depth-control due to the significant computational overhead and GPU resources required for these large-scale models, these results serve as a successful proof-of-concept. They demonstrate that our method is robustly applicable across various diffusion architectures. Qualitative examples for FLUX depth control are shown in Appendix Fig.~\ref{fig:flux_depth_qual}.

\textbf{Independent depth evaluation with DA-V2.}
Our primary depth evaluator is DPT-Large, while the reward loss uses DPT-Hybrid. To test whether control fidelity improvements generalize beyond the DPT family, we re-evaluate the same FLUX.1 generations using Depth Anything V2-Large (DA-V2)~\cite{yang2024depth}, which is used neither during training nor for model selection (Table~\ref{tab:depth_anything_v2_main}). Alignment improves RMSE under both evaluators: alone, it reduces RMSE from $33.1$ to $32.1$ with DPT-Large and from $33.9$ to $28.9$ with DA-V2; combined with the reward loss, it reduces RMSE from $29.2$ to $27.9$ and from $32.4$ to $31.5$, respectively. Notably, alignment alone achieves the best DA-V2 score, with a larger gain than reward-only training ($5.0$ vs. $1.5$ RMSE), confirming that its benefits extend to an independent estimator.

\begin{table}[h!]
\centering
\caption{\textbf{Independent depth evaluation on FLUX.1.} RMSE ($\downarrow$) for the same generations under DPT-Large and Depth Anything V2-Large. A/R denote alignment/reward; R+A is InnerControl.}
\label{tab:depth_anything_v2_main}
{
\begin{tabular}{lcccc}
\toprule
\textbf{Estimator} & \textbf{Base} & \textbf{R} & \textbf{A} & \textbf{R + A (Ours)} \\
\midrule
DPT-Large   & 33.1 & 29.2 & 32.1 & \textbf{27.9} \\
DA-V2-Large & 33.9 & 32.4 & \textbf{28.9} & 31.5 \\
\bottomrule
\end{tabular}}
\vspace{-6pt}
\end{table}

%% file: sec/ablations.tex
\subsection{Ablation}

\begin{table}[h!]
    \centering
    \caption{Ablation study on the Depth Map control task from the MultiGen-20M benchmark, analyzing the influence of the number of timesteps over which reward and alignment losses are applied.}    
    \label{tab:ablation_depth_steps}
    \resizebox{\columnwidth}{!}{%
    \begin{tabular}{l c c c c c}
        \toprule
        \textbf{Method} & \textbf{Steps} & \textbf{Steps Reward} & \textbf{RMSE $\downarrow$} & \textbf{FID $\downarrow$} & \textbf{CLIP $\uparrow$}\\
        \midrule
        \multicolumn{6}{l}{\textit{Guidance scale = 7.5}}\\
        \midrule
        ControlNet              & 0 & 0 & 33.95 & 18.61  &  \textbf{32.175} \\
        ControlNet              & 920 & 0 & 32.80 & 18.55            & 32.05 \\
        ControlNet              & 0 & 920 & \textbf{25.70} & 22.43 & 31.43 \\
        ControlNet              & 0 & 400 & 27.63 & 18.59 & 32.00 \\
        ControlNet              & 0 & 200 & 29.66 & 18.51 & \underline{32.11} \\
        ControlNet              & 200 & 200 & 28.93 & 18.32 & 32.06 \\
        ControlNet              & 400 & 200 & 28.56 & 18.57 & 32.00 \\
        ControlNet              & 600 & 200 & 28.41 & 18.52 & 31.99 \\
        ControlNet              & 920 & 200 & 27.50 & \textbf{18.22} & 31.92 \\
        ControlNet              & 920 & 400 & \underline{26.09} & \underline{18.29} & 32.00 \\
        \bottomrule
    \end{tabular}}
\vspace{-8pt}
\end{table}

\textbf{Alignment steps ablation.} We conducted an ablation study to analyze how the application of alignment and reward losses across different diffusion denoising steps affects performance (Table~\ref{tab:ablation_depth_steps}).  Specifically, we trained ControlNet models with alignment ($\mathcal{L}_{\text{alignment}}$) and reward ($\mathcal{L}_{\text{reward}}$) losses applied to varying subsets of denoising steps. All models were initialized from open-source ControlNet weights for depth estimation and trained under identical conditions (same seed, iteration count, and optimizer settings). Our experiments show that integrating alignment loss into both ControlNet and ControlNet++ training pipelines improves control alignment (RMSE) and image quality (FID). However, using alignment loss alone is less effective than reward loss. 

We further investigated the impact of extending the alignment loss to different numbers of denoising steps. Unlike reward loss, applying alignment loss to early steps does not increase FID. For this reason, in our main experiments, we apply alignment loss across 920 denoising steps. 

\textbf{Separate alignment loss.} We ablate the alignment loss on depth-control across SD1.5, SDXL, and FLUX(Table~\ref{tab:combined_ablation_narrow}). Used alone, the alignment loss gives modest RMSE gains over Base without hurting FID, confirming it as a safe, lightweight control signal. The reward loss alone yields much larger RMSE drops but often worsens FID (e.g., SDXL and FLUX; on SD1.5, R matches Base FID). Combining both (R+A) achieves the lowest RMSE across all backbones while also mitigating FID degradation relative to reward-only on FLUX, slightly improving FID on SD1.5, and matching reward-only FID within seed noise on SDXL, showing that alignment regularizes and complements reward-based training -- the core idea behind InnerControl. 

\begin{table}[ht]
\centering
\caption{Ablation study. R -- reward, A -- alignment, R+A -- proposed InnerControl}
\label{tab:combined_ablation_narrow}
\resizebox{\columnwidth}{!}{%
\begin{tabular}{ll cccc}
\toprule
\textbf{Model} & \textbf{Metric} & \textbf{Base} & \textbf{R} & \textbf{A}   & \textbf{R+A (InnerControl)} \\
\midrule
SD1.5 & RMSE$\downarrow$ & 34.0          & 27.6 & 32.8          & \textbf{26.1} \\
      & FID$\downarrow$  & 18.6          & 18.6 & 18.6          & \textbf{18.3} \\
\midrule
SDXL  & RMSE$\downarrow$ & 34.9          & 30.2 & 33.8          & \textbf{28.3} \\
      & FID$\downarrow$  & \textbf{15.5} & 17.0 & 15.6          & 17.1 \\
\midrule
FLUX  & RMSE$\downarrow$ & 33.1          & 29.2 & 32.1          & \textbf{27.9} \\
      & FID$\downarrow$  & 16.0          & 23.6 & \textbf{15.8} &  22.5 \\
\bottomrule
\end{tabular}
}
\vspace{-8pt}
\end{table}

\textbf{Pipeline-Agnostic Improvement.}
Beyond generalizing across diffusion backbones, the alignment loss is also agnostic to the underlying training pipeline. Table~\ref{tab:ctrlu_comparison} reports results on SD1.5 when adding our alignment loss on top of three distinct pipelines: the base ControlNet, reward-based fine-tuning, and CTRL-U. In every case, adding alignment improves RMSE without degrading FID. These results show that our alignment loss is not tied to a specific training procedure: it acts as a general-purpose plug-in that consistently complements and strengthens any existing pipeline. 

\begin{table}[ht]
\centering
\caption{Pipeline agnostic comparison (SD1.5)}
\label{tab:ctrlu_comparison}
\setlength{\tabcolsep}{4pt}
\begin{tabular}{ll cc}
\toprule
\textbf{Model} & \textbf{Metric} & \textbf{Base} & \textbf{Base + Alignment} \\
\midrule
ControlNet & RMSE$\downarrow$ & 34.0 & \textbf{32.8} \\
           & FID$\downarrow$  & \textbf{18.6} & \textbf{18.6} \\
\midrule
ControlNet++     & RMSE$\downarrow$ & 27.6 & \textbf{26.1} \\
           & FID$\downarrow$  & 18.6 &\textbf{18.3} \\
\midrule
CTRL-U     & RMSE$\downarrow$ & 26.5 & \textbf{25.4} \\
           & FID$\downarrow$  & 18.7 & \textbf{18.5} \\
\bottomrule
\end{tabular}
\vspace{-10pt}
\end{table}

%% file: sec/conclusion.tex


\section{Conclusion}
In this work, we introduce a training objective that aligns input controls with intermediate diffusion features across task-dependent ranges of the sampling trajectory. Unlike reward-based methods that primarily operate at late denoising steps, our lightweight probe enables alignment at earlier stages where spatial structure emerges. Our method is \textbf{architecture-agnostic} and \textbf{training-pipeline-agnostic}. Experiments on four control benchmarks demonstrate improved controllability while typically preserving perceptual quality, with FID remaining competitive across settings. These findings highlight the value of early-to-late alignment and the potential of our approach for further research.

%% file: sec/appendix.tex
\section{Implementation Details}
\label{sec:Implementation_details}
\paragraph{Training details.}

We utilize the same $\alpha$ for the reward loss weights as suggested in ControlNet++: $0.5$ for the depth and segmentation control, $1$ for the HED and $10$ for LineArt control with a timestep threshold of 200 steps for LineArt and segmentation tasks and $400$ steps for depth and HED control. For the alignment loss, training details are presented in the Table~\ref{tab:details}. Datasets information is shown in Table~\ref{tab:dataset}.  The training was conducted on $8$ H100 GPUs and took around $6$ hours. 
\begin{table}[ht!]
    \centering
    \caption{Details about some training parameters and reward models. $\text{ControlNet}^*$ denotes utilizing the same model to extract signal as ControlNet~\cite{zhang2023adding}}
    \label{tab:details}
    \resizebox{\linewidth}{!}{
    \begin{tabular}{l c c c c} 
        \toprule
        & \textbf{Segmentation Mask} & \textbf{Depth Map} & \textbf{Hed Edge} & \textbf{LineArt Edge} \\
        \midrule
        Reward Model (RM) & UperNet-R50 & DPT-Hybrid & $\text{ControlNet}^*$ & $\text{ControlNet}^*$ \\
        \midrule
        RM Performance & ADE20K(mIoU): 42.05 & NYU(AbsRel): 8.69 & - &  - \\
        \midrule
        Evaluation Model (EM) & Mask2Former & DPT-Large & $\text{ControlNet}^*$ & $\text{ControlNet}^*$ \\
        \midrule
        EM Performance & ADE20K(mIoU): 56.01 & NYU(AbsRel): 8.32 & - & - \\
        \midrule
        Reward Loss & CrossEntropy Loss & MSE Loss & MSE Loss & MSE Loss \\
        Loss Weight $\alpha$ & 0.5 & 0.5 & 1.0 & 10.0 \\
        Steps threshold & 200 & 400 & 400 & 200 \\
        \midrule
        Alignment Loss & CrossEntropy Loss & MSE Loss & Sparse MSE Loss & MSE Loss \\
        Loss Weight $\beta$ & 0.05 & 1.0 & 0.1 & 2.0 \\
        Steps threshold & [450, 980] & [0, 920] & [0, 800] & [0, 700] \\
        \bottomrule
    \end{tabular}
    }
\end{table}

\begin{table}[h!]
    \centering
    \caption{Datasets and evaluation details for explored tasks. $\uparrow$ denotes higher is better, $\downarrow$ -- lower is better.}
    \label{tab:dataset}
    \resizebox{\linewidth}{!}{
    \begin{tabular}{l c c c c}
        \toprule
        & \textbf{Segmentation Mask} & \textbf{Depth Map} & \textbf{HED Edge} & \textbf{LineArt Edge} \\
        \midrule
        Dataset & ADE20K~\cite{zhou2019semantic,zhou2017scene} & MultiGen20M~\cite{qin2023unicontrol}  & MultiGen20M~\cite{qin2023unicontrol} & MultiGen20M~\cite{qin2023unicontrol} \\
        \midrule
        Training Samples & 20,210 & 2,560,000 & 2,560,000 & 2,560,000 \\
        \midrule
        Evaluation Samples & 2,000 & 5,000 & 5,000 & 5,000 \\
        \midrule
        Evaluation Metric & mIoU $\uparrow$ & RMSE $\downarrow$ & SSIM $\uparrow$ & SSIM $\uparrow$ \\
        \bottomrule
    \end{tabular}
    }
\end{table}

We additionally estimate throughput and memory usage for ControlNet++ and our pipeline on $4$ H100 with batch size $8$. The results are presented in Table~\ref{tab:performance_metrics}.

For the Transformer-based depth control experiments, we keep the same training objective but apply the reward loss from steps 0 to 200, and the alignment loss from steps 400 to 920. In these runs, the alignment probe follows a compact, DiTF-inspired feature-selection strategy, while the rest of the training pipeline remains unchanged.

Our codebase is based on the implementation in HuggingFace’s Diffusers~\cite{von2022diffusers}. 

\paragraph{Reward models details.} We additionally provide information about reward models in Table~\ref{tab:details}. Following ControlNet++~\cite{li2024controlnet++} and CTRL-U~\cite{zhang2024ctrl} we utilize a slightly weaker model as the reward model for depth estimation and segmentation control training and a stronger model for evaluation. For HED and LineArt we use the same models as proposed in ControlNet~\cite{zhang2023adding}.


\begin{table}[h]
    \centering
    \caption{Component Performance Metrics}
    \label{tab:performance_metrics}
    \resizebox{\columnwidth}{!}{%
    \begin{tabular}{lcc}
        \toprule
        \textbf{Component} & \textbf{Time per iteration} & \textbf{Peak Memory} \\
        & & \textbf{Delta per GPU} \\
        \midrule
        Reward-only step & $0.15$ s & - \\
        Alignment-only step & $0.01$ s & - \\
        Full training iteration (rewarding) & $0.47$ s & $41$ Gb \\
        Full training iteration (ours) & $0.46$ s & $44$ Gb \\
        \bottomrule
    \end{tabular}%
    }
\end{table}

\section{Additional ablations} We conducted additional experiments to explore the impact of the alignment loss coefficient $\beta$ and the specific timesteps during which the alignment loss is applied for the depth control and segmentation control tasks. The results are presented in Tables~\ref{tab:ablation_depth} and~\ref{tab:ablation_segmentation}. These results indicate that, for segmentation, the alignment loss is most effective when applied over the later part of the denoising trajectory.

\begin{table}[h!]
    \centering
    \caption{Ablations on different $\beta$ values for alignment loss in the depth control task.}
    \label{tab:ablation_depth}
    \begin{tabular}{l c c c c c}
        \hline
         $\beta$ & loss  & g.s. & RMSE & FID & CLIP \\
        \hline
        0.5 & MSE & 7.5 & 26.09& 17.82 & 31.72 \\
        \hline
        1 & MSE & 7.5 & 26.09 & 18.29 & 32.00 \\
        \hline
        2 & MSE & 7.5 & 26.21 & 18.24 & 31.82 \\
        \hline
        5 & MSE & 7.5  & 25.98 & 18.08 & 31.68 \\
         \hline
    \end{tabular}
\end{table}


\begin{table}[h!]
    \centering
    \caption{Ablations on different $\beta$ values and the steps at which alignment loss is applied in the segmentation control task.}
    \label{tab:ablation_segmentation}
    \resizebox{\columnwidth}{!}{%
    \begin{tabular}{lcccccc}
        \toprule
        $\beta$ & steps & loss & g.s. & mIoU & FID & CLIP \\
        \midrule
        0.05 & $[450, 980]$ & CrossEntropy & 7.5 & 40.22 & 37.7 & 30.2  \\
        0.1  & $[450, 980]$ & CrossEntropy & 7.5 & 41.68 & 40.9 & 29.90 \\
        0.1  & $[400, 980]$ & CrossEntropy & 7.5 & 41.68 & 43.0 & 29.91 \\
        0.1  & $[450, 920]$ & CrossEntropy & 7.5 & 40.20 & 41.1 & 26.25 \\
        \bottomrule
    \end{tabular}%
    }
\end{table}


\begin{table*}[h!]
    \centering
    \caption{Per-condition semantic alignment measured by CLIP-score ($\uparrow$).}
    \label{tab:clip_per_condition}
    \begin{tabular}{@{}l c c c c c@{}}
        \toprule
        \multirow{2}{*}{\textbf{Method}} &
        \multirow{2}{*}{\textbf{T2I Model}} &
        \multicolumn{1}{c}{\textbf{Hed}} &
        \multicolumn{1}{c}{\textbf{LineArt}} &
        \multicolumn{1}{c}{\textbf{Depth}} &
        \multicolumn{1}{c}{\textbf{Segmentation}} \\
        & & CLIP $\uparrow$ & CLIP $\uparrow$ & CLIP $\uparrow$ & CLIP $\uparrow$ \\
        \midrule
        \multicolumn{6}{l}{\textit{Guidance scale = 7.5}}\\
        \midrule
        T2I‑Adapter                  & SD1.5 & — & — & 31.46 & — \\
        Gligen                       & SD1.4 & — & — & 31.48 & — \\
        Uni‑ControlNet               & SD1.5 & 31.94 & — & 31.66 & — \\
        UniControl                   & SD1.5 & 32.02 & — & 31.68 & — \\
        ControlNet                   & SD1.5 & 31.46 & 31.26 & 32.05 & 30.6 \\
        ControlNet++                 & SD1.5 & 32.05 & 31.87 & 32.0  & 30.9 \\
        Ctrl‑U                       & SD1.5 & 32.05 & 31.88 & 31.9  & 31.2 \\
        \textbf{InnerControl} (Ours) & SD1.5 & 32.05 & 31.78 & 32.0  & 30.2 \\
        \bottomrule
    \end{tabular}
\end{table*}

\section{Timestep schedules} Intervals for alignment loss application were determined through empirical grid search. Rather than searching blindly over all intervals, we first evaluated probe quality across diffusion timesteps (e.g., cross-entropy for segmentation and MSE for depth) and compared it against the corresponding reward model signal. For segmentation, probe predictions became unreliable in the final denoising steps (Fig.~\ref{fig:reward_readout}), which motivated us to exclude these late timesteps from the alignment loss. Guided by this observation, we then performed a coarse grid search over start/end points to identify the interval that improved controllability without degrading image quality.

\begin{figure}[h!]
    \centering
    \includegraphics[width=0.8\linewidth]{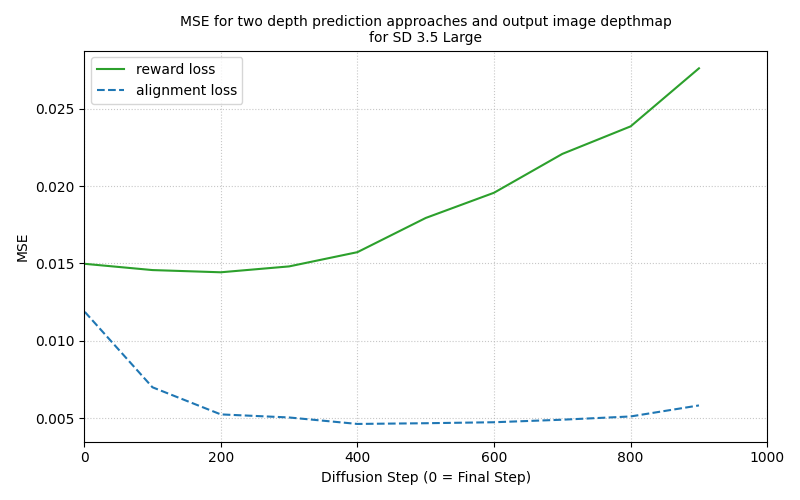}
    \caption{Reward-model and alignment-probe MSE across diffusion timesteps for depth control (SD3.5). The reward signal is strongest at low-noise steps, while the alignment probe remains stable over a wider step range, motivating the split schedule used in the experiments.}
    \label{fig:reward_vs_readout_sd35}
\end{figure}

For the Transformer-based depth experiments, we additionally compare reward model MSE and alignment MSE across diffusion timesteps. The reward signal is most reliable at the low-noise end of the trajectory, whereas the alignment probe remains stable over a broader early-to-mid step range. This motivates the split schedule used for such backbone depth runs (Fig.~\ref{fig:reward_vs_readout_sd35}).

\section{Additional results}
\subsection{Comparison of CLIP Score}
To estimate prompt alignment, we calculate CLIP-Score metrics, providing the results in Table~\ref{tab:clip_per_condition}. We calculate CLIP metrics for ControlNet++ and CTRL-U using our trained versions. We observe that while providing more aligned and quality images, we remain at the same CLIP score level in various setups.

\subsection{User study}
 We conducted a user study (Tab.~\ref{tab:user_study}) comparing InnerControl with Reward across control types. Annotators compared paired outputs (SD~1.5 for edges/segmentation, SDXL for depth) and selected the preferred result by control alignment, prompt fidelity, and aesthetic quality.

\begin{table}

  \caption{ \textbf{User study}. Results of pairwise comparisons on 100 randomly sampled validation image pairs. 12 participants took part in the study. Values denote the percentage of cases where InnerControl is preferred over Reward.
}
  \label{tab:user_study}
  \centering
  \begin{tabular}{lccc}
    \toprule
    \textbf{Question} & \textbf{Depth} & \textbf{Edges} &
    \textbf{Segment.}\\
    \midrule
    Prompt fidelity (Q1) &  55\% & 67\% & 50\% \\
    Control Alignment (Q2) & 65\% & 79\% & 72\% \\
    Aesthetic quality (Q3) & 63\% & 77\% & 85\% \\
    \bottomrule
  \end{tabular}
\end{table}


\section{Alignment models details}
For our work, we utilize the architecture for $\mathbb{H}(\cdot, t)$ from Readout Guidance~\cite{luo2024readout}. Following~\cite{luo2024readout}, we build an aggregation network that takes features from the UNet decoder, applies bottleneck layers~\cite{he2016deep} to standardize channel counts, and aggregates them through a learned weighted sum. Additionally, these models use pretrained timesteps embedding for model conditioning to make predictions on each diffusion step. 

The probe is trained on features of the base diffusion model, before and
independently of ControlNet fine-tuning, and is kept frozen throughout the subsequent alignment training: gradients from $\mathcal{L}_{\text{alignment}}$
propagate into the ControlNet only. At inference the probe is discarded entirely, so it adds no cost to sampling.

\begin{figure}[h!]
    \centering
    \includegraphics[width=0.7\linewidth]{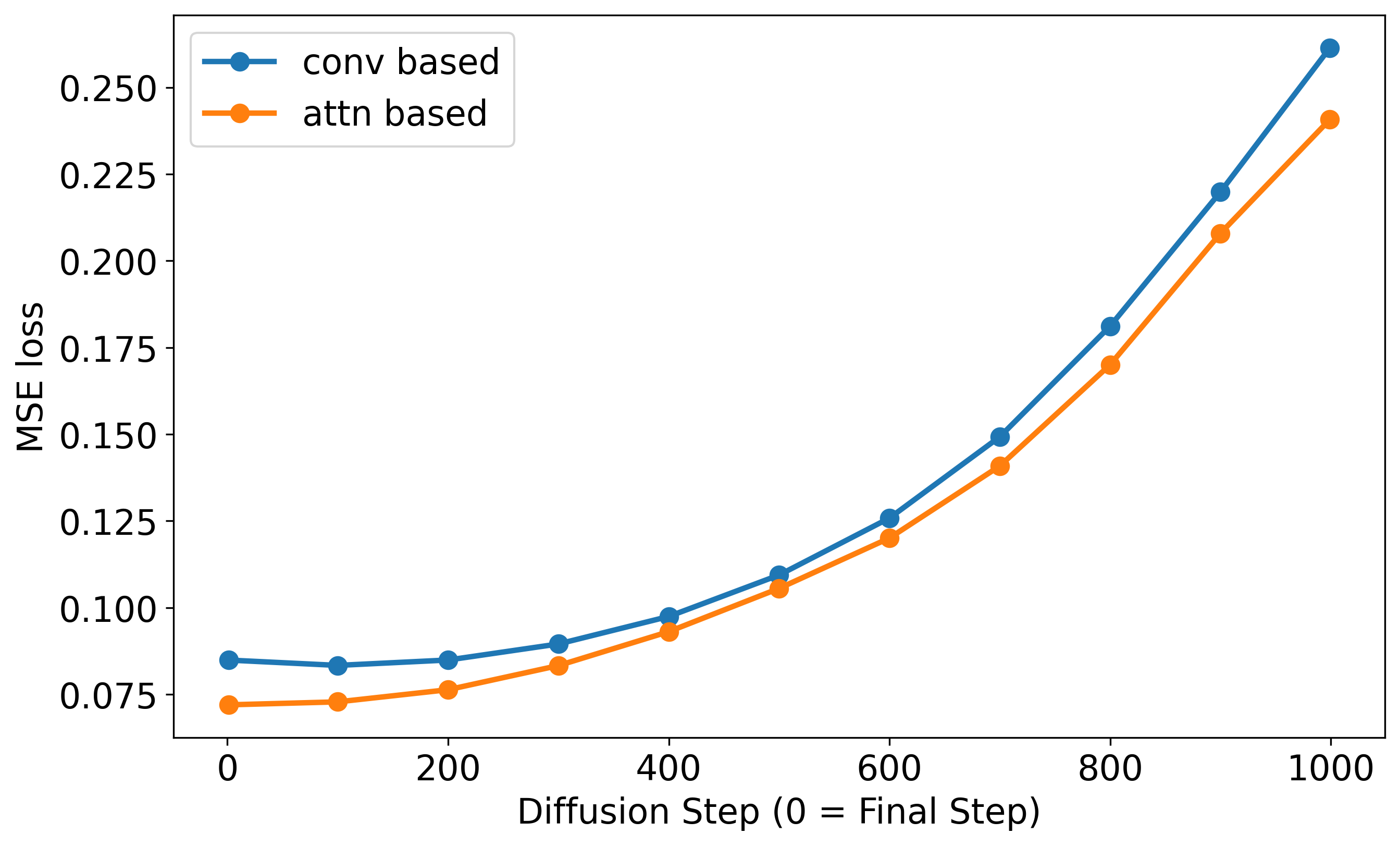}
    \caption{Quality comparison for attention-based and convolution-based predictions for depth maps extraction task}
    \label{fig:attn_vs_conv}
\end{figure}

For the depth-control task, we obtain slightly better estimates by using self-attention features from the UNet decoder instead of convolutional features (Fig.~\ref{fig:attn_vs_conv}). For edge controls such as HED and LineArt, we resize decoder features to $128 \times 128$ and add a transposed-convolution output head, which improves recovery of thin structures. For segmentation, we keep the baseline probe design as it already demonstrates strong performance.

\begin{table}[ht]
\centering
\caption{Comparison of Alignment Model Parameters across different architectures.}
\label{tab:model_parameters}
\begin{tabular}{lc}
\toprule
\textbf{Backbone} & \textbf{Num. of Parameters} \\
\midrule
UNet         & 8.50M \\
UNet (Unified) & 9.49M \\
SD3.5 (fast) & 8.10M \\
SD3.5 (max)  & 79.8M \\
FLUX         & 9.90M \\
\bottomrule
\end{tabular}
\end{table}
For Transformer-based backbones, we keep the same timestep-conditioned aggregation design but cannot rely on UNet-style decoder readouts, since these architectures are organized as a single sequence of Transformer blocks. Instead, following DiTF~\cite{gan2025unleashing}, we extract modulated Transformer-block features and rearrange the selected patch tokens into 2D feature maps before aggregation. This allows to preserve the lightweight nature of the probe while adapting it to the representation structure of SD3.5 and FLUX.

These models contain around 8.5M parameters (Table~\ref{tab:model_parameters}). For comparison, Stable Diffusion 1.5 has approximately 980M parameters and FLUX.1 has about 12B parameters, so the probes are roughly two to three orders of magnitude smaller than the main denoiser.

The models were trained for 10000 steps for edge tasks and 5000 for depth and segmentation using the Adam optimizer with a batch size of 8, while the learning rate was kept fixed at 1e-3 throughout training. For training, we employed several datasets: ADE20K~\cite{zhou2019semantic, zhou2017scene} for the segmentation task, PascalVOC~\cite{everingham2010pascal} to annotate custom controls for HED Edges with the HED model and LineArt Edges using extractors proposed in ControlNet~\cite{zhang2023adding}), and DPT~\cite{ranftl2021vision} for depth estimation. 
The main loss function used across all models was mean squared error (MSE), except for the segmentation task, where cross-entropy loss was applied.

\begin{figure*}[h!]
    \centering
    \includegraphics[width=0.95\linewidth]{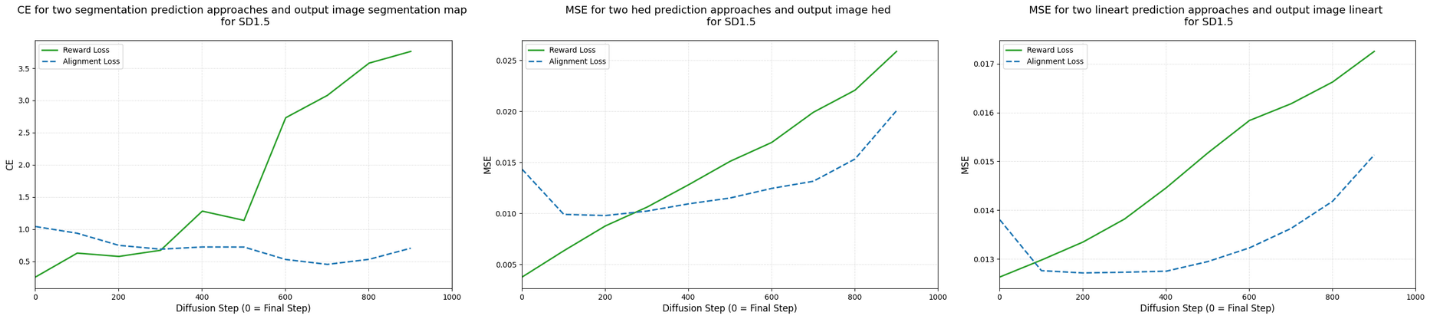}
    \caption{Comparison of signals estimated from one-step prediction and intermediate features for segmentation, HED, and LineArt control tasks.}
    \label{fig:reward_readout}
\end{figure*}

\begin{figure*}[h!]
    \centering
    \includegraphics[width=0.8\linewidth]{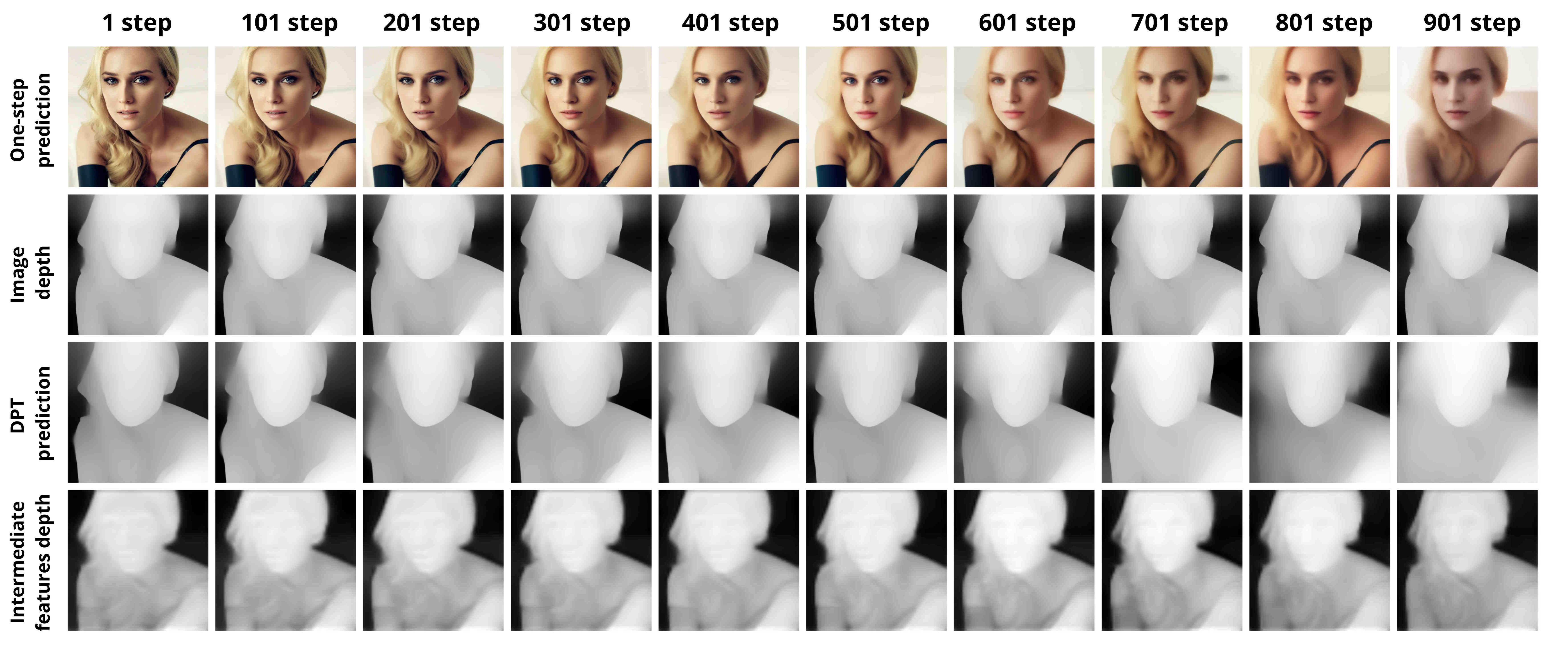}
    \caption{Visualization of one-step prediction, estimated depth of the generated image, corresponding DPT depth estimation, and depth extracted from intermediate features.}
    \label{fig:inner_depth}
\end{figure*}

\begin{figure*}[h!]
    \centering
    \includegraphics[width=0.8\linewidth]{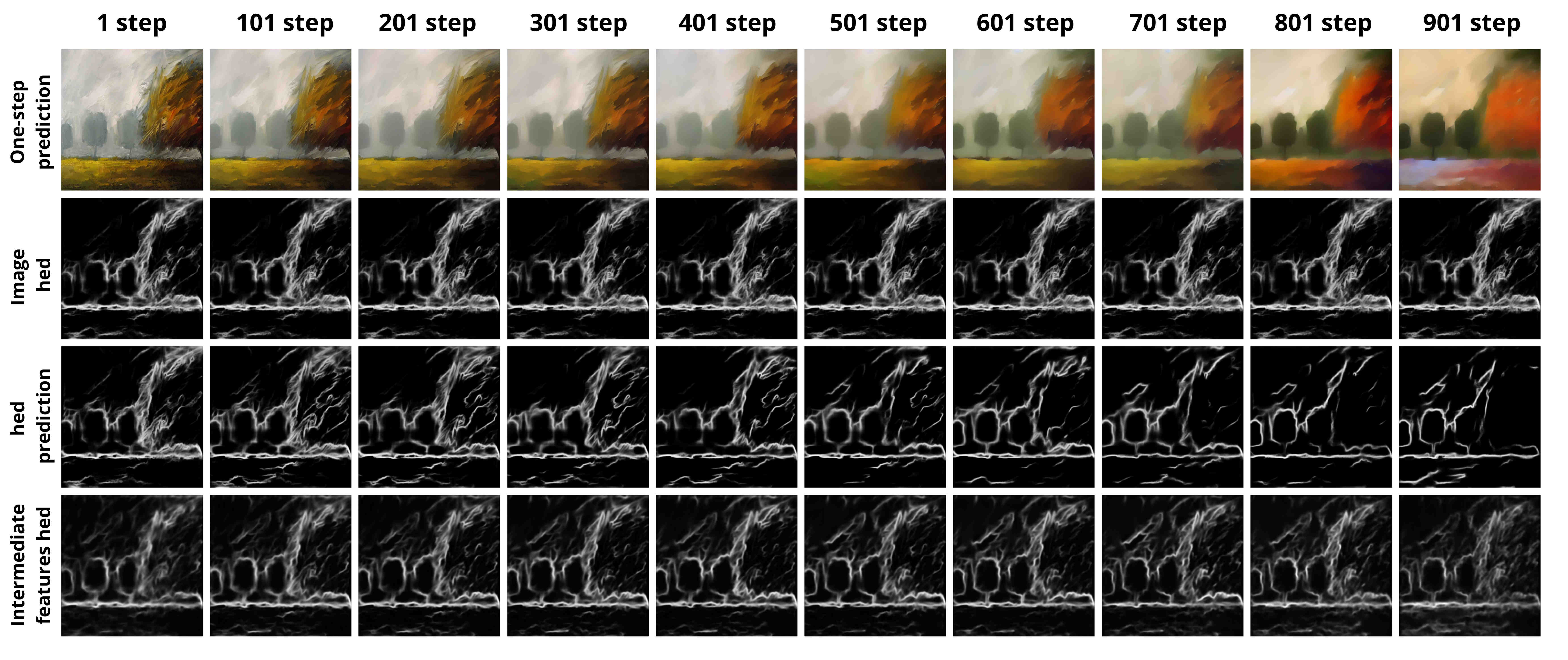}
    \caption{Visualization of one-step prediction, estimated HED of the generated image, corresponding HED estimation, and HED extracted from intermediate features.}
    \label{fig:inner_hed}
\end{figure*}

\begin{figure*}[h!]
    \centering
    \includegraphics[width=0.8\linewidth]{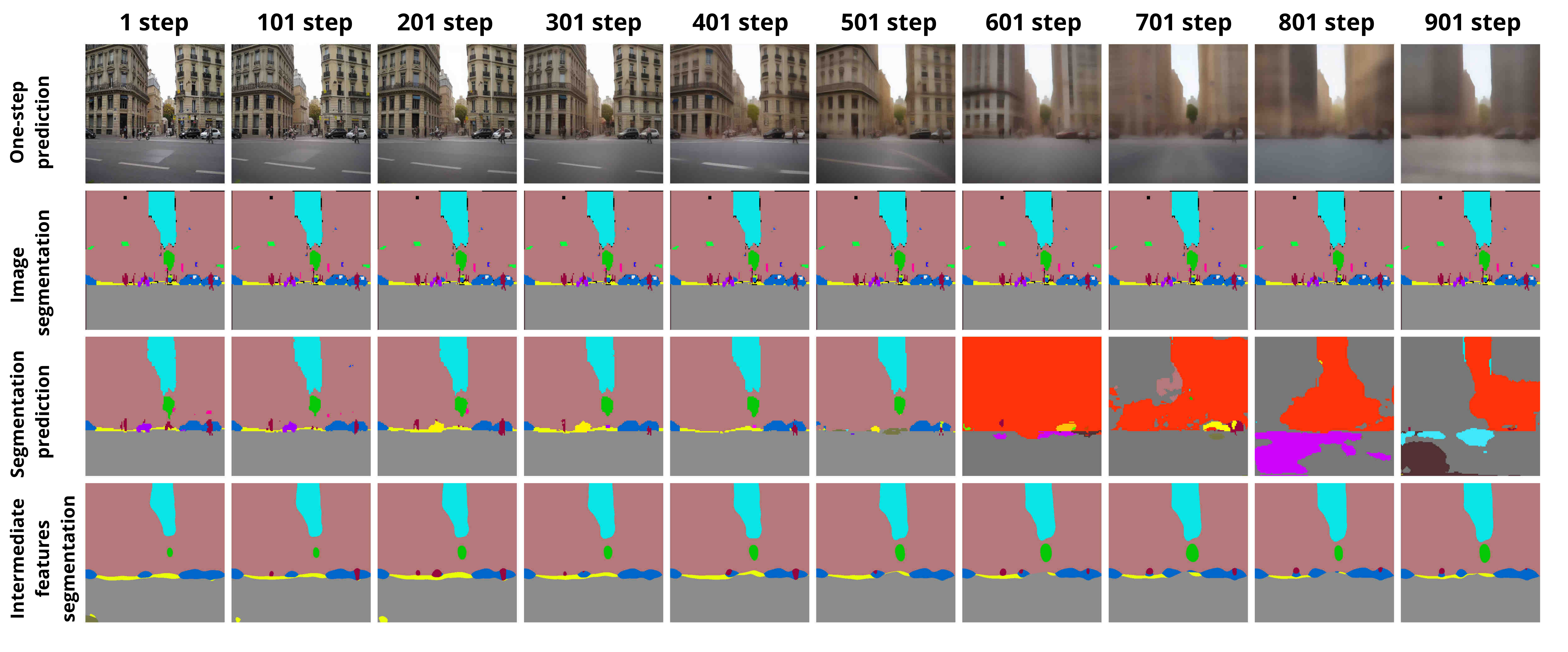}
    \caption{Visualization of one-step prediction, estimated segmentation mask of the generated image, corresponding segmentation maps, and segmentation maps extracted from intermediate features.}
    \label{fig:inner_seg}
\end{figure*}

\begin{figure*}[h!]
    \centering
    \includegraphics[width=0.8\linewidth]{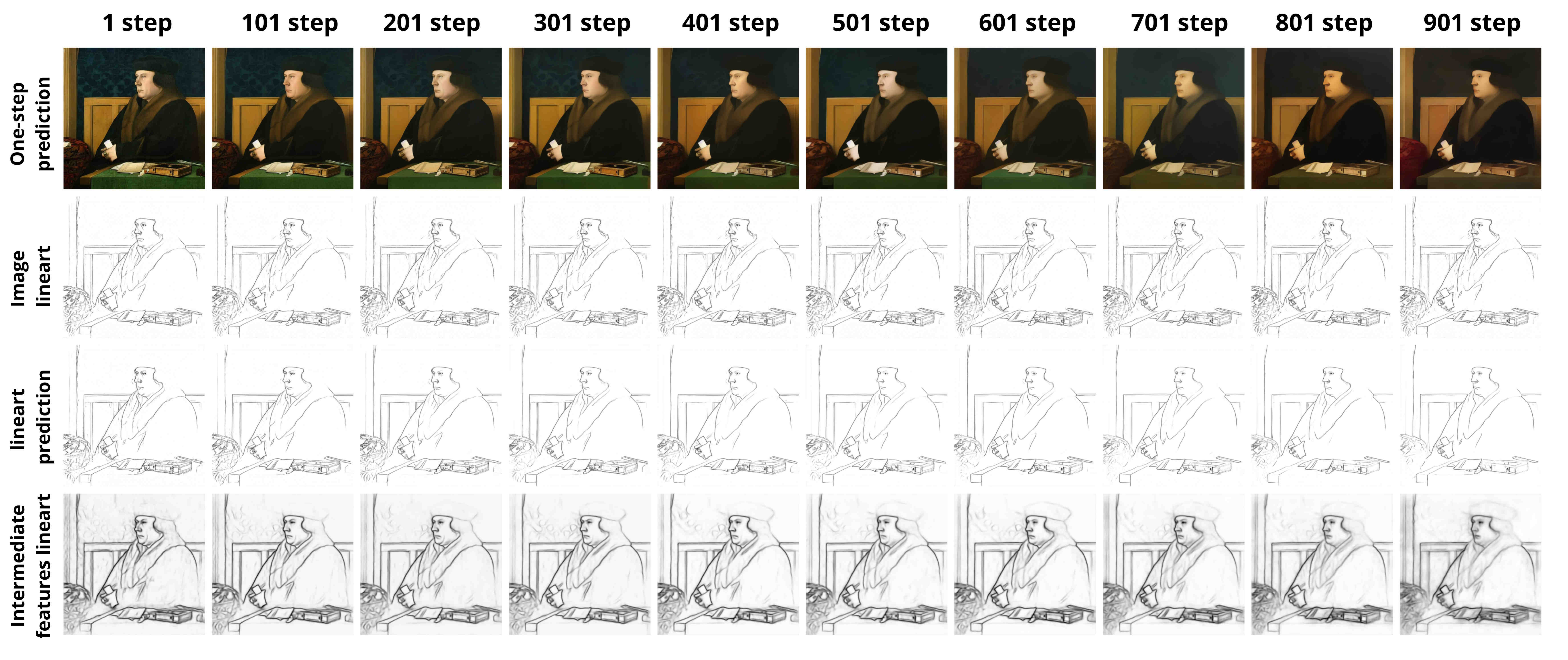}
    \caption{Visualization of one-step prediction, estimated LineArt of the generated image, corresponding LineArt estimation, and LineArt extracted from intermediate features.}
    \label{fig:inner_lineart}
\end{figure*}

For a more detailed look, we provide additional visualizations of the intermediate control signals, including the extracted depth, segmentation maps, and edges (HED and LineArt), in Figs.~\ref{fig:inner_depth},~\ref{fig:inner_hed},~\ref{fig:inner_seg}, and~\ref{fig:inner_lineart}. Additionally we calculate Cross-Entropy (CE) for segmentation and MSE for edge tasks to compare reward modeling and signal from intermediate features in Fig.~\ref{fig:reward_readout}. These figures reveal that standard discriminative models struggle to extract precise control information from blurred images, especially for depth and segmentation estimation tasks.

\begin{figure}[h!]
    \centering
    \includegraphics[width=0.7\linewidth]{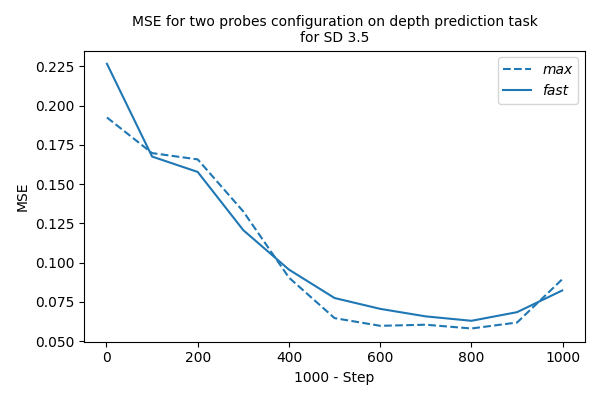}
    \caption{Probe MSE across diffusion timesteps for two SD3.5 probe configurations. The \emph{fast} configuration, which uses only five selected Transformer layers, remains close to the \emph{max} configuration that aggregates all valid layers.}
    \label{fig:probe_max_vs_fast}
\end{figure}

\paragraph{Transformer probe selection.} Unlike the UNet backbone, the Transformer denoisers do not expose a separate encoder/decoder split, so we cannot directly reuse decoder readouts. Following DiTF~\cite{gan2025unleashing}, we compare final probe variants: a \emph{max} configuration that aggregates all valid Transformer blocks and a \emph{fast} configuration that keeps features only from five layers with the highest semantic correspondence as shown in Table 7 of~\cite{gan2025unleashing}. Both use the same automatic top-1 channel-discard rule, which gives a small improvement in probe loss. The resulting fast probes remain compact -- 8.1M parameters for SD3.5 and 9.9M for FLUX -- which is comparable to the 8.5M SD1.5 probe (Table~\ref{tab:model_parameters}) and negligible relative to the roughly 8B and 12B denoisers. Figure~\ref{fig:probe_max_vs_fast} shows that the fast configuration, which is 10 times smaller than the max configuration, tracks it closely across diffusion timesteps. 

\begin{figure*}[h!]
    \centering
    \includegraphics[width=0.97\linewidth]{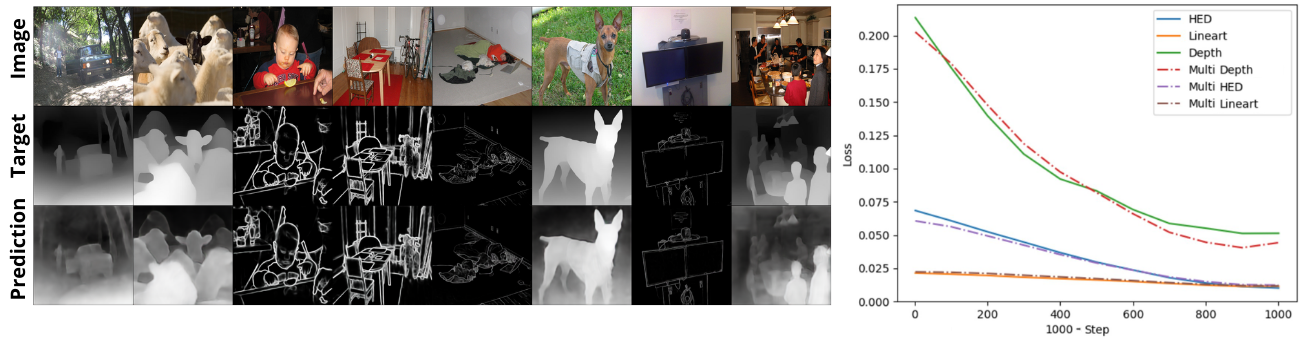}
    \caption{Quantitative and Qualitative Comparison of Models. Results comparing the performance of the unified model against individual, single-task models on three diverse visual tasks: Depth estimation, Holistically-Nested Edge Detection (HED), and LineArt extraction.}
    \label{fig:unified_probe}
\end{figure*}

\paragraph{Unified alignment model.} While training separate probes is already light\-weight, we additionally trained a unified multi-task probe to address pipeline-complexity concerns directly. This model has 9.49M parameters, uses a shared bottleneck with task-specific heads, and is trained by randomly sampling the task at each iteration and applying the corresponding supervised loss. Because it learns depth, HED, and LineArt jointly, training was extended to 30,000 iterations. The unified model reaches performance comparable to separate task-specific probes (Fig.~\ref{fig:unified_probe}). Evaluation is conducted on UNet generation using PascalVOC-based controls.

\begin{figure}[h]
    \centering
    \includegraphics[width=0.99\columnwidth]{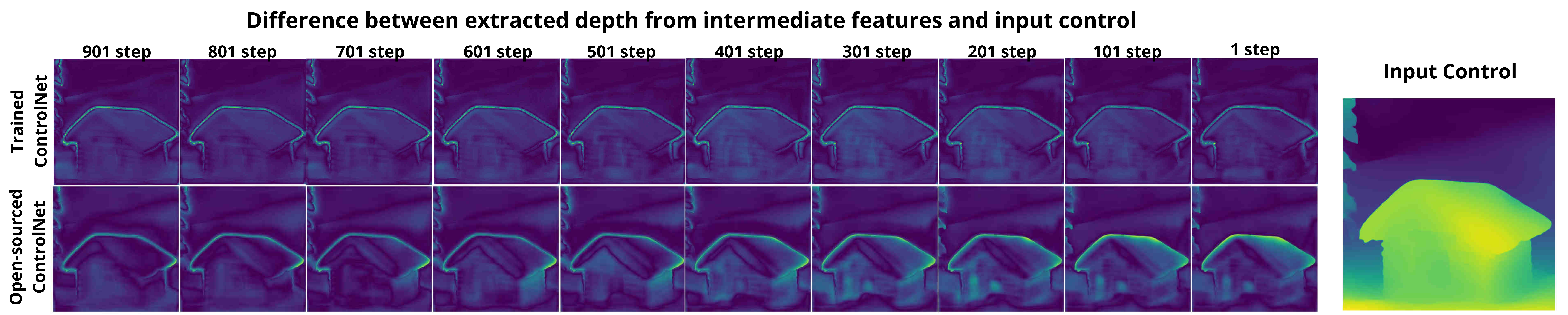}
    \caption{Visualization of difference between extracted signal from intermediate features and input control after our training applied (\textit{top}) and for standard ControlNet (\textit{bottom})}  
    \label{fig:diff}
\end{figure}

\paragraph{Intermediate features.} We compared extracted feature alignment with the input control for the depth estimation task in Fig.~\ref{fig:diff}. The top row illustrates that, after ControlNet training, extracted depth maps exhibit high correspondence with the control signal across different steps. This visualization highlights the effect of enforcing alignment throughout the sampling trajectory, improving both the extracted features and the final generated image.

\begin{figure*}[h!]
    \centering
    \includegraphics[width=0.95\linewidth]{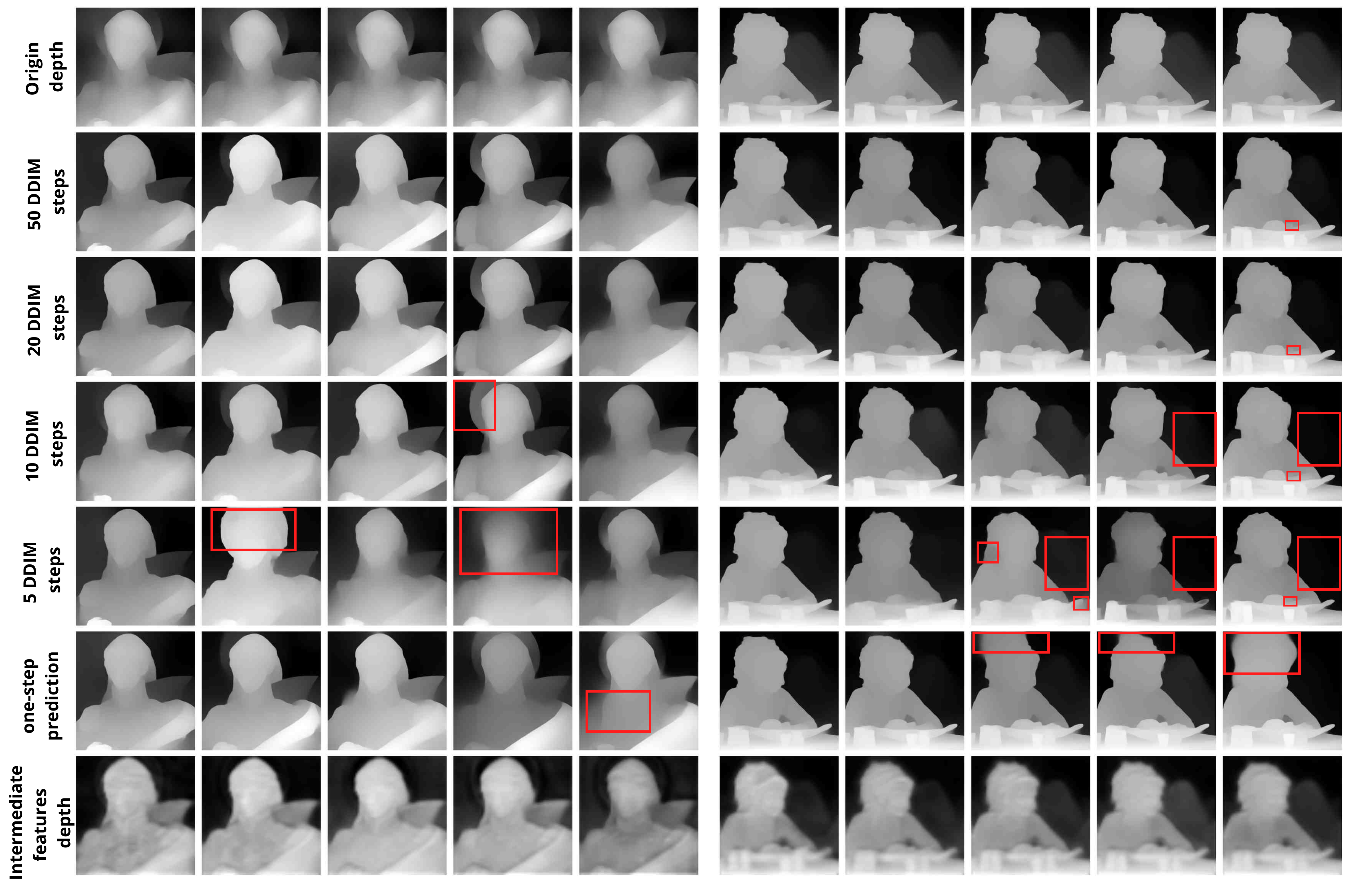}
    \caption{Visualization of various numbers of sampling steps for depth prediction. The red rectangles indicate artifacts detected in the depth predictions.}
    \label{fig:sampling}
\end{figure*}

\paragraph{Alternative to one-step prediction.} {While one-step prediction often produces noisy results on the early steps of generation, utilizing more sophisticated samplers like DDIM typically incurs high time and memory costs. We significantly optimized the DDIM sampling process by applying gradient checkpointing, and were able to run the full 50-step DDIM sampling with a peak memory usage of 66,265 MB ($\sim$66 GB) and a batch size 1 processing time of $\sim$3.83 seconds per loss computation. When reducing the number of steps to 25, we can fit a batch size of 2 on a single H100, with a peak memory of 73,298 MB ($\sim$73 GB) and a slightly higher per-batch time of $\sim$3.98 seconds. Reducing sampling steps more again leads to poor quality, see Fig.~\ref{fig:sampling}}

\begin{figure}[h]
    \centering
    \includegraphics[width=0.9\linewidth]{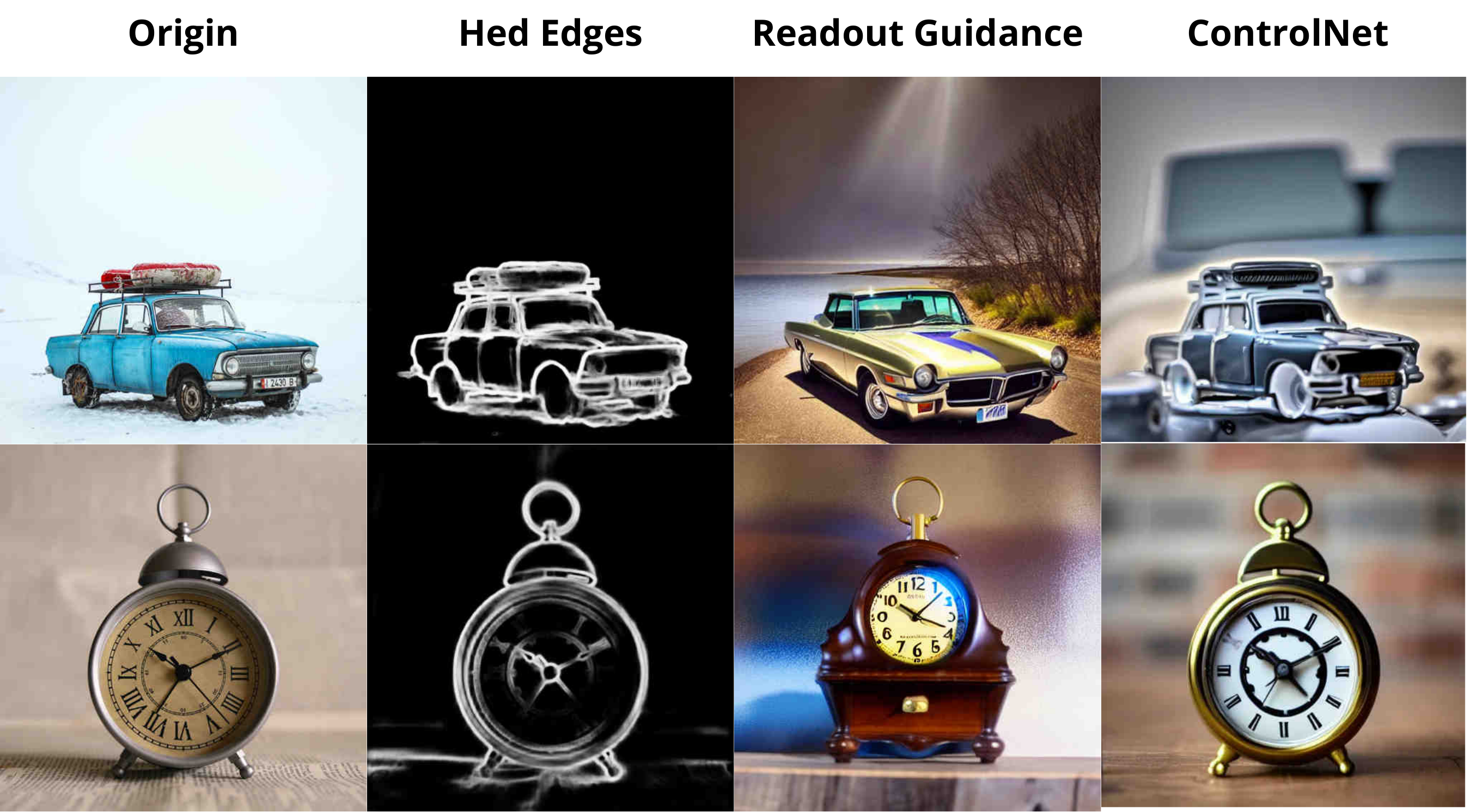}
    \caption{Application of small networks for guidance during inference.}
    \label{fig:rg}
\end{figure}

\section{Discussion}\label{discussion}

\paragraph{Utilizing small networks during inference.} The idea of using the probe at inference time for on-the-fly correction is closely related to Readout Guidance~\cite{luo2024readout}, which proposes augmenting Classifier-Free Guidance with an additional gradient for on-the-fly correction:
\begin{equation}
    \mathbf{\hat{\epsilon}}_t \leftarrow \mathbf{\epsilon}(x_t) + w \cdot \nabla_{x_t} d(r, f(x_t))
\end{equation}

where $f(\cdot)$ is the small prediction network (analogous to our probe), $r$ is the target control map, and $d(\cdot,\cdot)$ is a distance function.
We evaluated this approach using hed control. While theoretically appealing, this approach failed to improve controllability and severely degraded efficiency. The generated images exhibited poor alignment with the input control (e.g., HED), often failing to incorporate essential structural cues (Fig.~\ref{fig:rg}).

Another important limitation of such approach is inference time increasing. The on-the-fly gradient calculation, even when applied only during the first half of sampling, caused the inference time to increase. A single 50-step DDIM sample with SD 1.5 inflated from approximately 30 seconds to 1 minute 47 seconds.
Thus, our findings demonstrate that the lightweight prediction networks are highly effective when integrated into the training-time alignment loss (as shown by our core results), applying them for on-the-fly inference correction degrades efficiency and fails to improve output controllability. 

\section{Comparison with Native I2I Generation Models}
\label{app:i2i}
 
Recent image-generation models support image-to-image conditioning natively, without an auxiliary control branch. We evaluate two of them, FLUX.2~[klein] (9B)~\cite{flux2klein2025} and UniPic2 (2B)~\cite{wei2025unipic}, on the FLUX depth benchmark, using the same validation samples, prompts and depth conditions as Table~2. For both native-I2I baselines, we provide the MultiGen depth map itself as the sole reference/input image through the model's standard image-editing interface, together with the unchanged text prompt. Metrics follow the main paper.
 
\begin{table}[h]
\centering\small
\caption{Comparison with models supporting native I2I generation on the FLUX
depth benchmark. Depth conditions and prompts are identical across methods.}
\label{tab:i2i}
\setlength{\tabcolsep}{5pt}
\begin{tabular}{lccc}
\toprule
Method & RMSE $\downarrow$ & FID $\downarrow$ & CLIP $\uparrow$\\
\midrule
Klein   & 46.06 & 26.15 & 29.16\\
UniPic2 & 65.02 & 28.92 & 28.41\\
\midrule
ControlNet++ (FLUX.1) & 29.22 & 23.61 & $\mathbf{29.22}$ \\
\textbf{InnerControl (FLUX.1)} & $\mathbf{27.91}$ & $\mathbf{22.54}$ & 28.88 \\
\bottomrule
\end{tabular}
\end{table}
 
Both models are substantially less faithful to the geometric condition. This is expected: they are trained more for instruction-driven editing, where the conditioning image sets appearance and semantics rather than a dense geometric target, so they are not controlled baselines for spatial conditioning. We report the comparison as context -- at this level of precision native I2I does not replace an explicit
control branch, and an alignment objective of the kind proposed here remains applicable on top of whichever conditioning mechanism a backbone provides.
 
\paragraph{User study.}
Following the protocol of App.~D.2, $14$ participants compared $270$ randomly sampled pairs in this setting.

\begin{table}[h]
\centering\small
\caption{User study on the FLUX depth setting. Values denote the percentage of cases where InnerControl is preferred over the competing model.}
\label{tab:i2i_user}
\setlength{\tabcolsep}{5pt}
\begin{tabular}{lccc}
\toprule
Question & vs.\ ControlNet++ & vs.\ Klein & vs.\ UniPic2 \\
\midrule
Prompt \\fidelity (Q1)   & 94.9\% & 76.5\% & 84.1\% \\
Control \\alignment (Q2) & 88.0\% & 72.6\% & 91.5\% \\
Aesthetic \\quality (Q3) & 97.8\% & 31.1\% & 79.1\% \\
\bottomrule
\end{tabular}
\end{table}

The user study supports our quantitative results: participants consistently preferred InnerControl for prompt fidelity and control alignment against all three baselines, while Klein was favored for aesthetic quality. Although FLUX.2 [klein] was preferred aesthetically, it was noticeably less successful at following the provided prompts and control conditions. These results further indicate that the improved condition following of InnerControl is perceptually apparent to human evaluators.

\section{Limitations} The main limitation of our approach is the quality of small convolutional neural nets for signal estimation from intermediate features. Due to their small parameter count and shallow design, these models may struggle to predict fine-grained spatial details, such as thin edges. However, we emphasize that this limitation is not intrinsic to the method itself. Our framework may utilize any model capable of extracting a signal at each timestep. This opens a promising direction for developing a better model in future work.

\section{Ethics statement}

Our method builds upon pretrained text-to-image diffusion backbones, including Stable Diffusion and FLUX variants, and therefore inherits their potential biases and limitations.



\section{More visualizations}
\label{sec:Visualization}

\begin{figure*}[h!]
    \centering
    \includegraphics[width=0.7\linewidth]{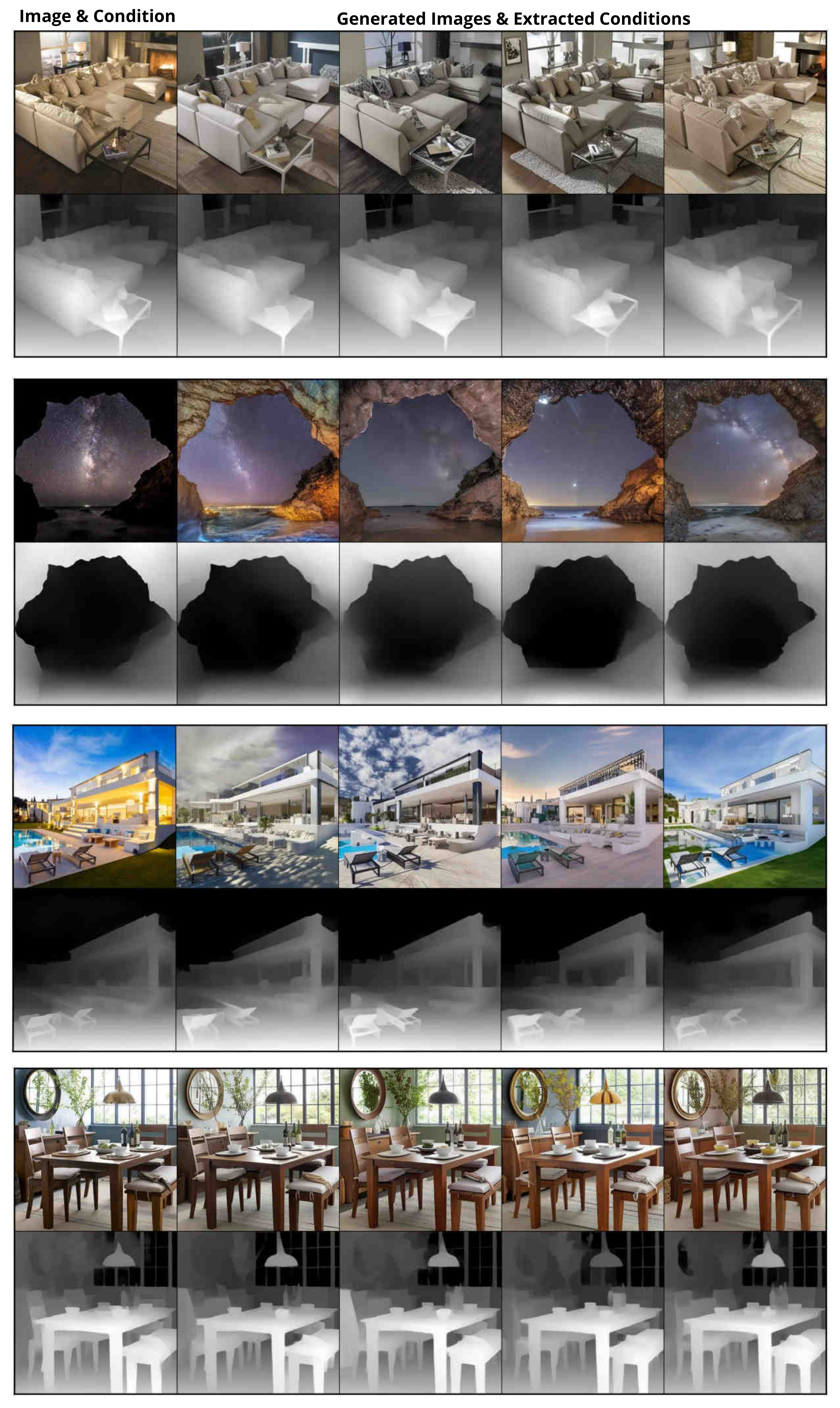}
    \caption{More visualizations for InnerControl (ours) method (depth maps)}
    \label{fig:appendix_depth}
\end{figure*}

\begin{figure*}[h!]
    \centering
    \includegraphics[width=0.7\linewidth]{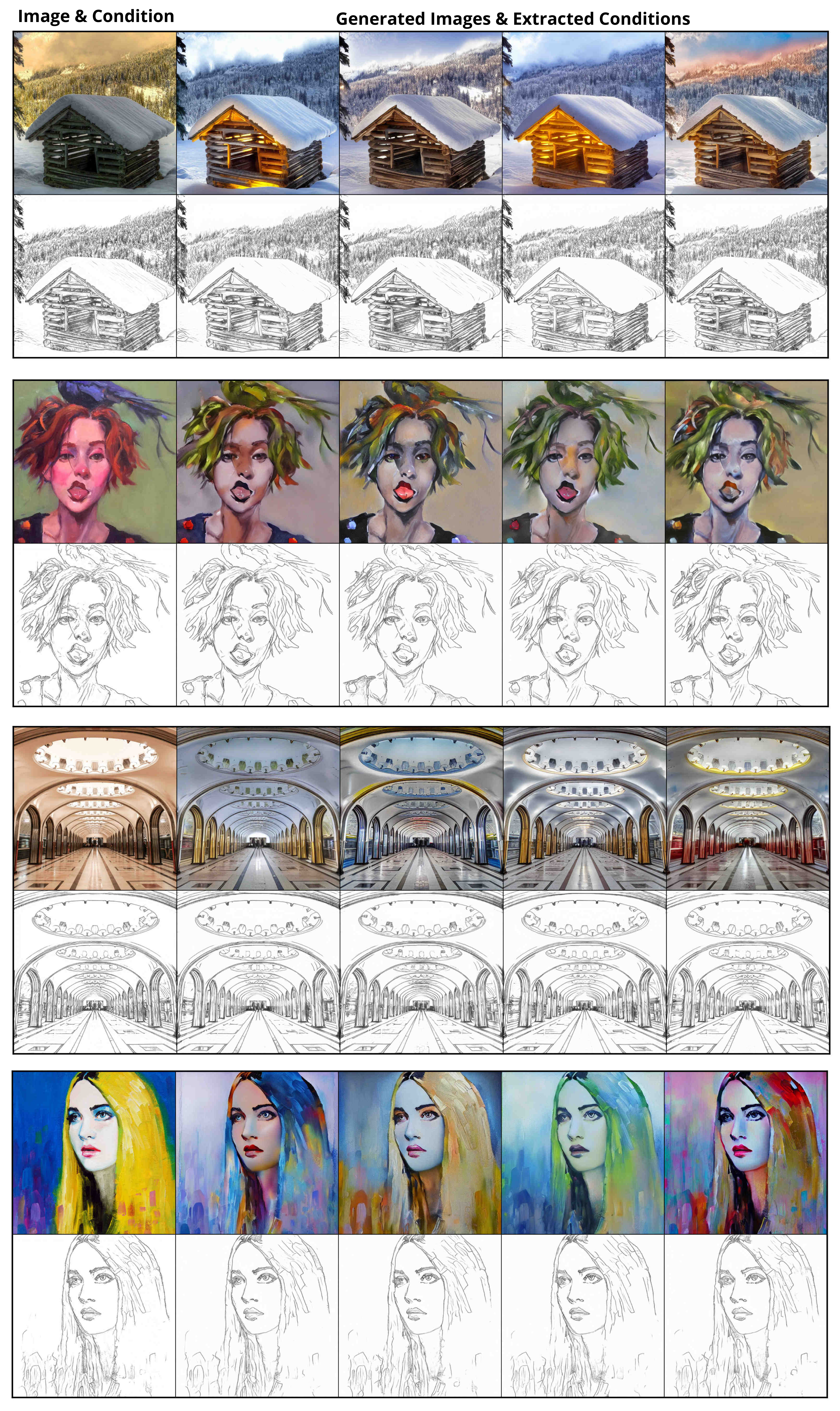}
    \caption{More visualizations for InnerControl (ours) method (LineArt)}
    \label{fig:appendix_lineart}
\end{figure*}

\begin{figure*}[h!]
    \centering
    \includegraphics[width=0.7\linewidth]{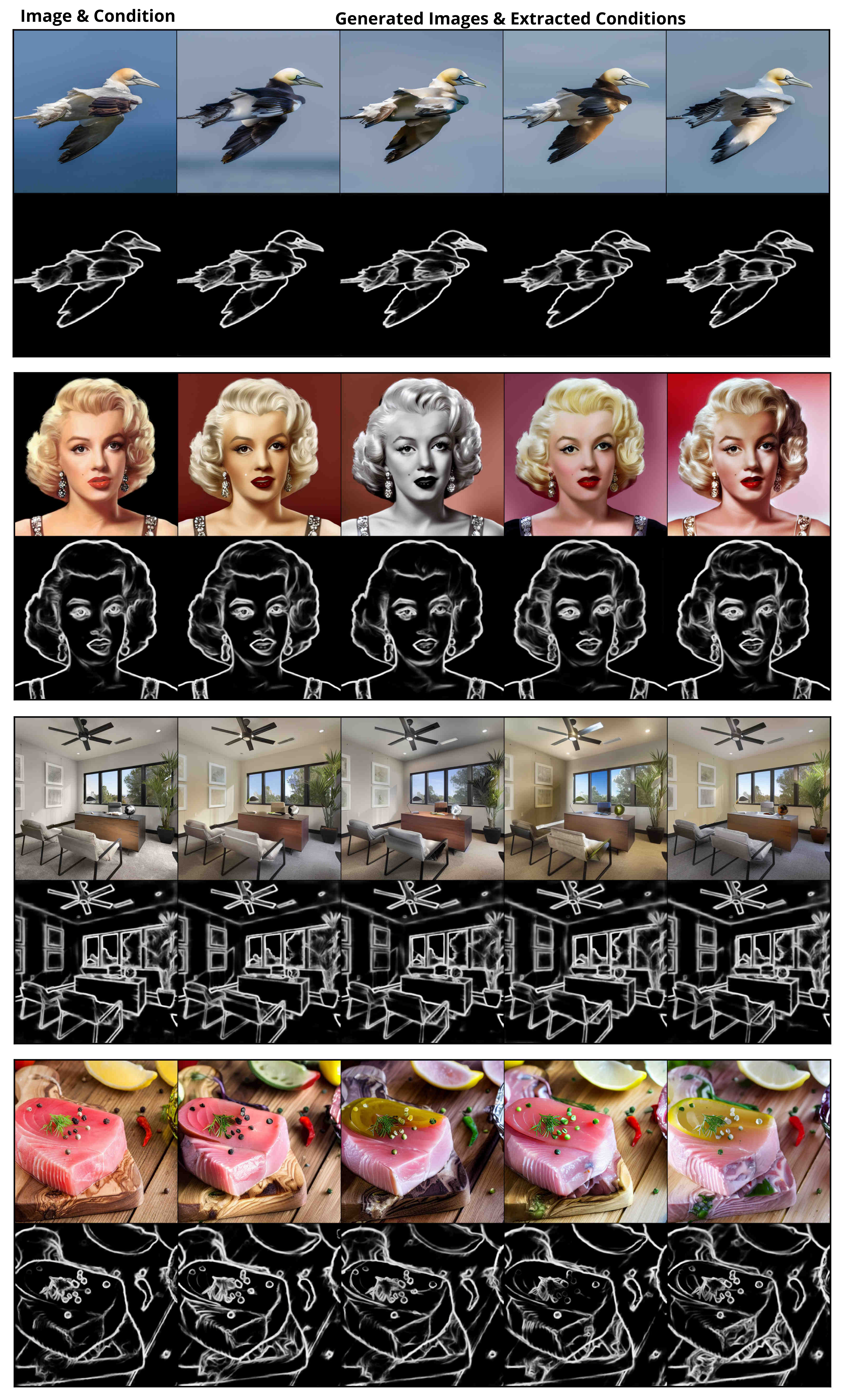}
    \caption{More visualizations for InnerControl (ours) method (HED)}
    \label{fig:appendix_hed}
\end{figure*}

\begin{figure*}[h!]
    \centering
    \includegraphics[width=0.7\linewidth]{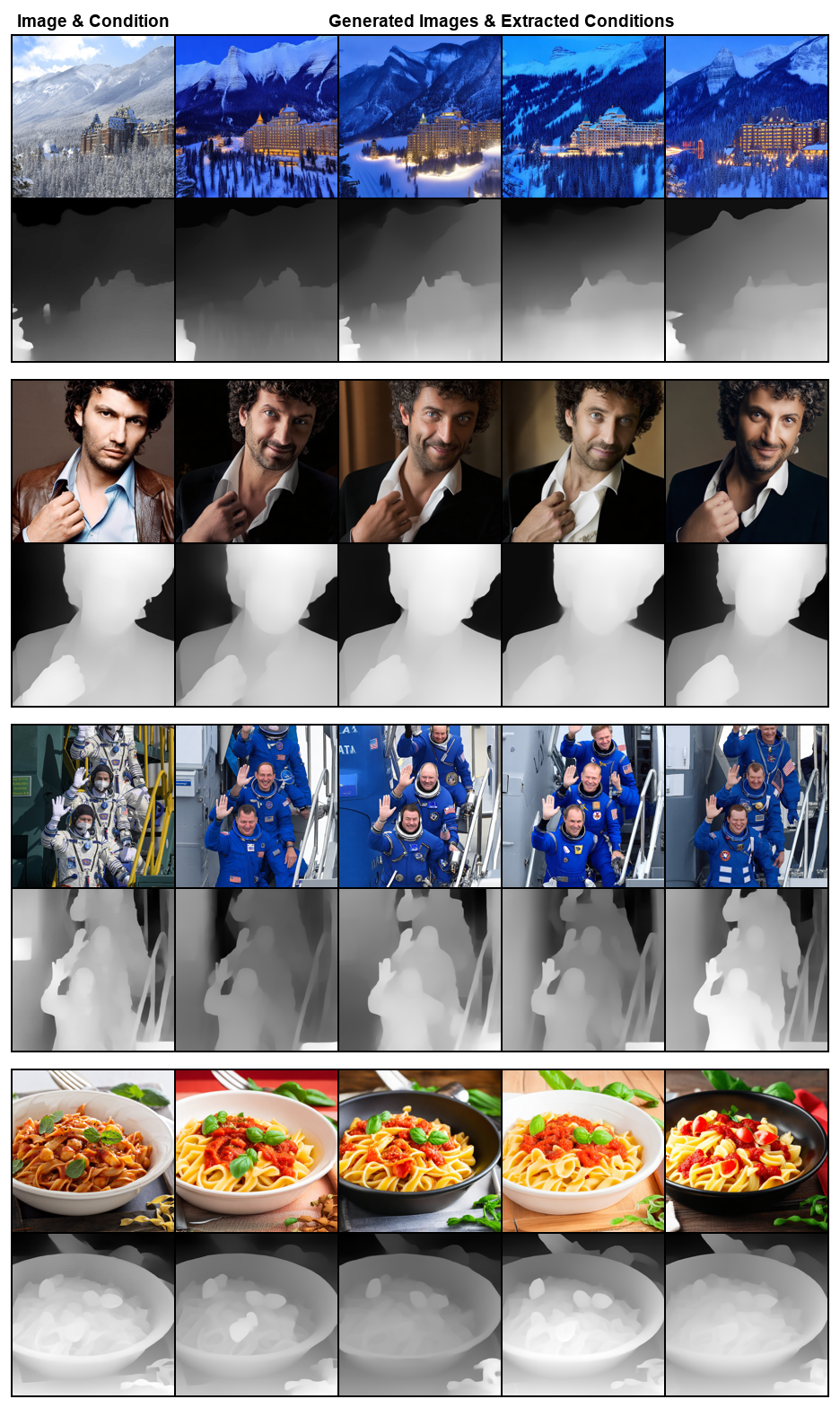}
    \caption{More visualizations for InnerControl (ours) method based on SD3.5 model (depth maps)}
    \label{fig:appendix_depth_sd35}
\end{figure*}

\begin{figure*}[h!]
    \centering
    \includegraphics[width=0.7\linewidth]{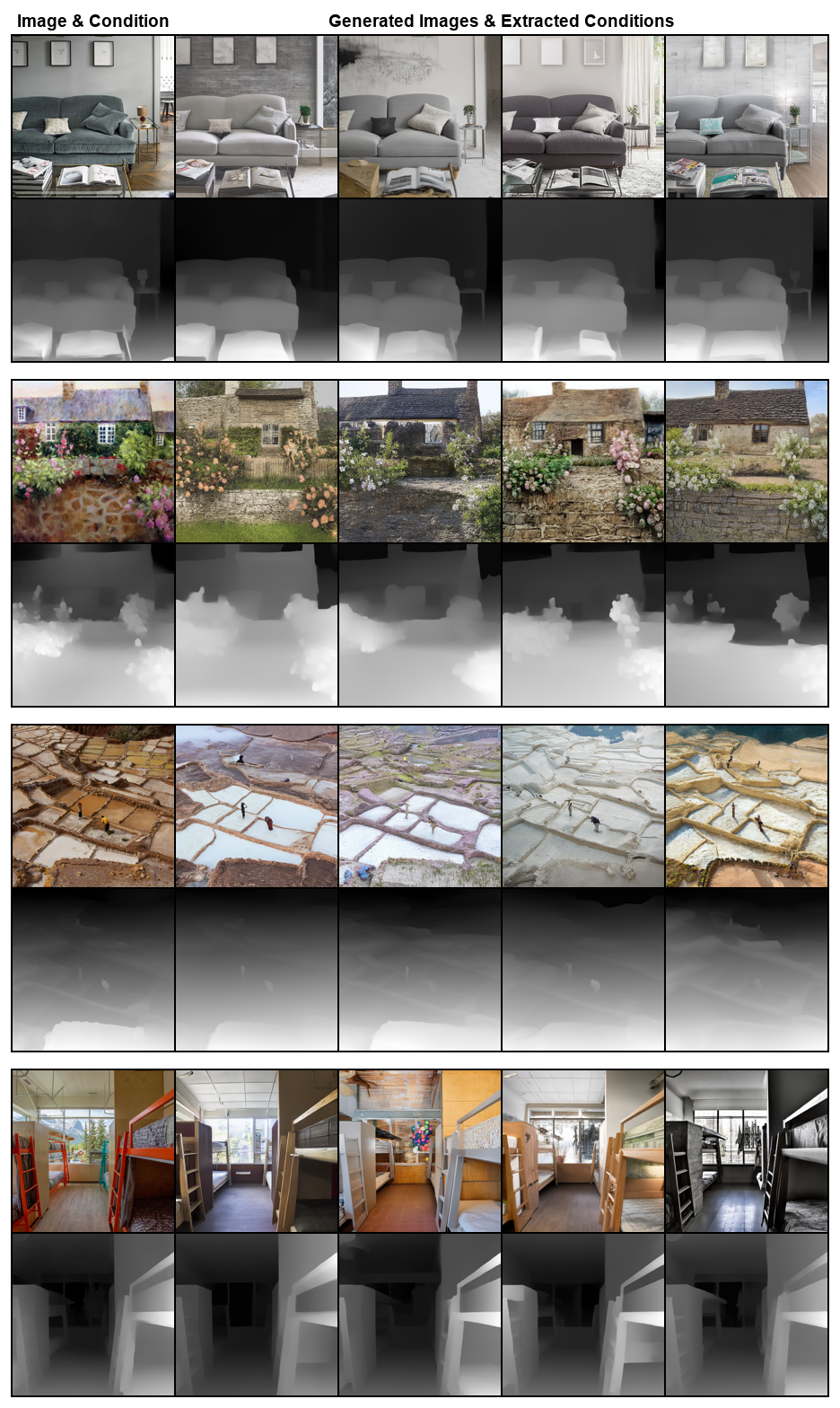}
    \caption{More visualizations for InnerControl (ours) method based on FLUX model (depth maps)}
    \label{fig:appendix_depth_flux}
\end{figure*}

We also provide additional qualitative examples for InnerControl generation. UNet-based results are shown in Figs.~\ref{fig:appendix_depth},~\ref{fig:appendix_lineart}, and~\ref{fig:appendix_hed}. Additional depth-control examples for the Transformer-based models are shown in Figs.~\ref{fig:appendix_depth_sd35} and~\ref{fig:appendix_depth_flux}.

\begin{figure*}[h!]
    \centering
    \includegraphics[width=0.95\linewidth]{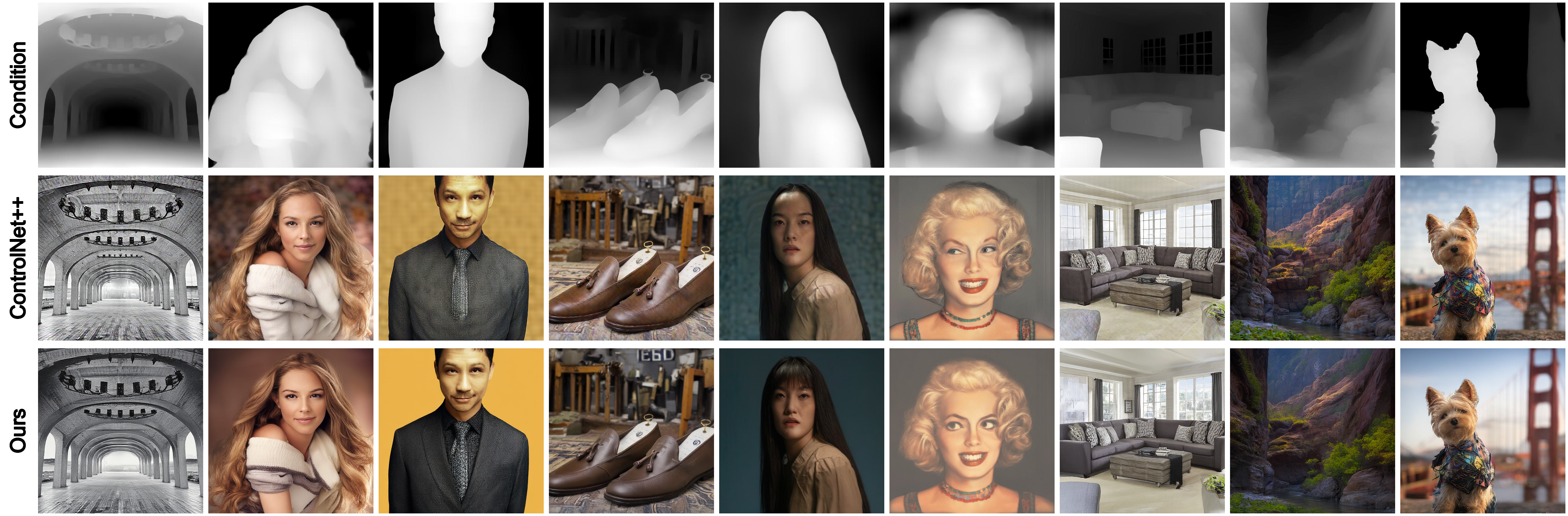}
    \caption{FLUX depth-control qualitative comparison (moved from main paper). For each sample: input depth condition, ControlNet++ and InnerControl output.}
    \label{fig:flux_depth_qual}
\end{figure*}